\documentclass[journal]{IEEEtran}

\usepackage{paralist,tikz,graphicx,amsmath, amsfonts,amssymb,amsthm,overpic, latexsym,epsfig,color,rotating,paralist,times,float,subcaption,algorithm,verbatim,enumitem,url}
\usetikzlibrary{fit,positioning}
\usepackage{tikz,pgfplots}

\newtheorem{theorem}            {Theorem}

\newtheorem*{definition*}{Definition}

\newcommand{\new}{N}
\newcommand{\known}{K}

\newcommand{\activ}{{\tt active}}
\newcommand{\passive}{{\tt passive}}

\newcommand{\threat}{{\tt threat}}
\newcommand{\high}{{\tt high}}
\newcommand{\low}{{\tt low}}

\newcommand{\horizon}{N}
\newcommand{\I}{\Pi}

\newcommand{\weather}{\text{review}}
\newcommand{\wealth}{w}

\newcommand{\Qfn}{Q}

\newcommand{\pdf}{p}
\newcommand{\cdf}{F}
\newcommand{\prob}{\mathbb{P}}
\newcommand{\E}                 {\Bbb{E}}

\renewcommand{\P}                 {\Bbb{P}}

\newcommand{\thbelief}{\eta}
\newcommand{\rewardp}{\reward_{\thbelief,2}}
\newcommand{\pdfp}{\pdf_{\thbelief}}

\newcommand{\obs}{y}

\newcommand{\G}{\mathcal{G}}

\newcommand{\beliefdim}{m}

\newcommand{\state}{x}
\newcommand{\generalstate}{s}
\newcommand{\statespace}{\mathcal{X}}
\newcommand{\physicalstatespace}{\mathcal{S}}
\newcommand{\obspace}{\mathcal{Y}}

\newcommand{\obsdim}{{Y}}

\newcommand{\fun}{\phi}

\newcommand{\oprob}{B}
\newcommand{\tp}{P}

\newcommand{\belief}{\pi}

\newcommand{\Belief}{\Pi}

\newcommand{\ole}{\stackrel{\text{defn}}{=}}

\newcommand{\lr}{\leq_r}

\newcommand{\Stop}{\mathcal{S}}

\newcommand{\filterd}{\sigma}

\newcommand{\filter}{T}
\newcommand{\filterg}{\bar{\filter}}
\newcommand{\filterdg}{\bar{\filterd}}

\newcommand{\globaltogo}{J}
\newcommand{\uglobaltogo}{\underline{J}}

\newcommand{\reals}{{\rm I\hspace{-.07cm}R}}

\newcommand{\cond}{\lambda}

\newcommand{\beq}{\begin{equation}}
\newcommand{\eeq}{\end{equation}}
\newcommand{\nn}{\nonumber}

\newcommand{\Y}{\mathbf{Y}}

\renewcommand{\th}{\theta}

\newcommand{\p}{\prime}

\newcommand{\ones}{\mathbf{1}}

\newcommand{\diag}{\textnormal{diag}}

\newcommand{\reward}{r}
\newcommand{\diffreward}{\Delta_\thbelief}

\newcommand{\Cost}{C}

\newcommand{\polp}[1]{\pol_{#1,\thbelief}^*}
\newcommand{\polpT}{\pol^*_{2,\filter(\belief,\obs)}}

\newcommand{\valueg}{\mathcal{\valuef}}

\newcommand{\action}{a}

\newcommand{\actionspace}{\,\mathcal{A}}

\def\param{{p}}

\newcommand{\policy}{\mu}

\newcommand{\optpolicy}{\policy^*}

\newcommand{\valuef}{V}

\newcommand{\valueb}{V}
\newcommand{\utilitytogo}{J}

\newcommand{\uV}{\underline{V}}

\newcommand{\ant}{\Psi}

\newcommand{\stopset}{\mathcal{S}}

\newcommand{\oprobg}{R^\belief}

\newcommand{\private}  {\eta}

\newcommand{\argmin}{\operatornamewithlimits{argmin}}
\newcommand{\argmax}{\operatornamewithlimits{argmax}}

\newcommand{\public}{\pi}
\newcommand{\changetime}{\tau^0}

\newcommand{\A}{\mathbb{A}}

\newcommand{\globalpolicy}{\phi}

\newcommand{\globalactionspace}{\mathcal{U}}
\newcommand{\globalaction}{u}

\newcommand{\dtime}{n}

\newcommand{\barray}{\begin{array}{ll}}
\newcommand{\earray}{\end{array}}

\newcommand{\physical}{s}
\newcommand{\psych}{z}

\newcommand{\pol}{\mu}
\newcommand{\vpol}{{\boldsymbol{\pol}}}

\newcommand{\utility}{V}

\newcommand{\umu}{\underline{\globalpolicy}}
\newcommand{\uQ}{\underline{Q}}

\newcommand{\uvalueg}{\underline{\valueg}}
\newcommand{\Mb}{M^\belief}
\newcommand{\red}[1] {\textcolor{red}{#1}}

\newcommand{\Ali}{A}

\newcommand{\wealthmax}{W_0}
\newcommand{\tv}{{\tt stock}}
\newcommand{\garden}{{\tt bond}}
\newcommand{\win}{{\tt good}}
\newcommand{\lose}{{\tt bad}}
\newcommand{\gardenreward}{\iota}
\newcommand{\acta}{{a}}
\newcommand{\actb}{\bar{a}}

\newcommand{\bet}{{\tt bet}}
\newcommand{\nobet}{{\overline{\tt bet}}}

\newcommand{\winbet}{\tt win}
\newcommand{\losebet}{\overline{\tt lose}}
\newcommand{\oprobparam}{\th}

  \usepackage{tikz}
\usetikzlibrary{patterns}

\tikzset{
    blockd/.style={rectangle, draw, line width=0.2mm, black, text width=10.8em,  text centered,
                 minimum height=2em},
    line/.style={draw, -latex}}      

\tikzset{
    blockb/.style={rectangle, draw, line width=0.2mm, black,  text centered,
                 minimum height=2em},
    line/.style={draw, -latex}}      

\tikzset{
    blocka/.style={rectangle, draw, line width=0.3mm, black, text width=6.2em, text centered,
                 minimum height=1em},
               line/.style={draw, -latex}}

             \tikzset{
    blockf/.style={rectangle, draw, line width=0.3mm, black, text width=13.2em, text centered,
                 minimum height=1em},
               line/.style={draw, -latex}}

\renewcommand\fbox{\fcolorbox{red}{white}}
 \newcommand{\blue}[1] {{#1}}
\begin{document}

\title{\huge Quickest Change Detection
  of Time Inconsistent Anticipatory Agents. 
  Human-Sensor and Cyber-Physical Systems}

\author{Vikram~Krishnamurthy, {\em Fellow IEEE}, 
\today
  \thanks{Vikram Krishnamurthy, School of Electrical and Computer Engineering, Cornell University.
    Email: vikramk@cornell.edu.  This research was supported by the U.S.\ Army Research Office under grant
    W911NF-19-1-0365.}}

\maketitle

\begin{abstract} In behavioral economics, human decision makers are modeled as anticipatory agents  that make decisions by taking into account the probability of future decisions (plans).  We  consider cyber-physical systems involving  the interaction between  anticipatory  agents and statistical detection. A sensing device records the decisions of  an anticipatory  agent.
  Given these decisions, how can the sensing device achieve quickest detection of  a change in the anticipatory system?
  From a decision theoretic point of view,  anticipatory  models are time inconsistent meaning that Bellman's principle of optimality does not hold. The appropriate formalism is the subgame Nash equilibrium.
We show that the interaction between anticipatory   agents  
and sequential quickest detection  results in unusual (nonconvex)  structure
of the quickest change detection policy.
Our methodology yields a useful  framework for
situation awareness systems and  
 anticipatory  human decision makers interacting with 
  sequential detectors.
\end{abstract}

\subsection*{Glossary of Symbols}
\begin{tabular}{cl}
    \multicolumn{2}{l}{{\bf Anticipatory agent.} Sec.\ref{sec:clmodel} and \ref{sec:structure}} \\
  $\physical_1,\physical_2$  & physical state\\
  $\psych_1, \psych_2$ & psychological state  (\ref{eq:psych}), (\ref{eq:ant}) \\
  $\action_{1},\action_{2}$ &  actions (\ref{eq:pol12}) \\
  $\optpolicy_1,\optpolicy_2$ & Nash equilibrium  policy  (\ref{eq:period1}), (\ref{eq:period2}) \\
    $\valueb_1(\cdot)$, $\valueb_2(\cdot)$  & value function \\
  \multicolumn{2}{l}{{\bf Quickest detection.} Sec.\ref{sec:qdnew}} \\
  $\dtime$ & discrete time $\dtime$ (also agent $\dtime$) \\
  $\state_\dtime$ & jump state (for quickest detection)\\
  $\tp$ & transition matrix of $\{\state_\dtime, \dtime \geq 0\}$  (\ref{eq:tp}) \\
  $f,d$ & false alarm and delay penalty parameters \\
   \multicolumn{2}{l}{{\bf Anticipatory agents acting sequentially.}
 Sec.\ref{sec:qdnew}} \\
    $\physical_\dtime$ & physical state\\
   $\psych_\dtime$  & psychological state                 \\
       $\action_{\dtime_1},\action_{\dtime_2}$ & local decision maker's actions
  \\
    $\private_\dtime$ & private belief of local decision maker $\dtime$  (\ref{eq:public}) \\
  $\optpolicy_{\dtime,1},\optpolicy_{\dtime,2}$ & Nash equilibrium policy   (\ref{eq:period1}), (\ref{eq:period2}) \\
  $\obs_\dtime$ & private  observation of $\state_n$  at time $\dtime$\\
$  \oprob_{\state_\dtime,\obs_\dtime}$ & observation likelihood $\pdf(\obs_\dtime|\state_\dtime)$  (\ref{eq:oprob}) \\
  $\filter(\belief,\obs)$ & private belief update   (\ref{eq:privateupdate}) \\
    $\filterd(\belief,\obs)$ & normalization measure for private belief \\
  \multicolumn{2}{l}{{\bf Global Decision maker.} Sec.\ref{sec:qdnew} and Sec.\ref{sec:structure2}}\\ 
   $\globalaction_\dtime $&  action at time $\dtime \in \{1 \text{(stop)}, 2 \text{(cont)}\}$
  \\
   $\globalpolicy^*(\belief,\physical)$ &  optimal policy for quickest detection \\
  $\belief_\dtime$ & public belief at $\dtime$  (\ref{eq:public}) \\
  $\oprobg_{\state,\action}(\physical) $ & action likelihood $\pdf(\action| \state, \belief,\physical) $ (\ref{eq:actionprob}), (\ref{eq:oprobgcompute}) \\
  $\filterg(\belief,\action,\physical) $ & public belief update (\ref{eq:pubupdate})  \\
  $\filterdg(\belief,\action,\physical) $ & normalization measure for public belief \\
  $\valueg(\belief,\physical)$ & value function  \\
$  \Cost(\belief,\globalaction)$  & costs incurred in quickest detection

\end{tabular}

\noindent {\bf Keywords}  Time inconsistency, anticipatory decision making, subgame Nash equilibrium, quickest change detection, change blindness, Blackwell dominance, multi-threshold policy

\noindent {\bf Acknowledgment}. The author is grateful to   Professor Andrew Caplin, Department of Economics,  NYU for numerous  suggestions and discussions regarding his  influential paper \cite{CL01}.

\section{Introduction} \label{sec:intro}
`Cognitive sensing' is widely used in signal processing, but lacks the important property of  anticipatory decision making.
{\em An anticipatory agent makes decisions by taking to account the probability of  future decisions}. This crucial property is studied in behavioral economics involving human decision makers
and yields remarkable behavior such as time inconsistency as  discussed below.

This paper is an early step in understanding the interaction between statistical detection and  behavioral economics models. Signal processing and behavioral economics are mature areas; yet their intersection, namely cyber-physical systems involving interaction of human decision makers  with sensing based detection is relatively unexplored.
The main question we address  is: {\em If multiple anticipatory decision makers interact sequentially (or a single anticipatory agent acts repeatedly), how can a global decision maker use these anticipatory decisions to achieve optimal sequential change detection?}


\begin{figure}[h] 
  \centering
  \fbox{ \begin{minipage}{8 cm} \mbox{} \vspace{-0.5cm}  \\
      \begin{tikzpicture} [node distance =2.5cm and 2cm, auto]
                  \tikzset{every node}=[footnotesize=\small]
                  \node [blocka] (l2) {Anticipatory \\ agent 1};
                  \node [blocka,right of=l2] (l3) {Anticipatory agent 2 };
                   \node [blocka,right of=l3] (l4) {Anticipatory agent 3 };
                   \node [right of=l4,node distance=2cm] (cdots)[draw=none]{$\cdots$};
                   
    \node [blockd,below of=l3,node distance=1.75cm] (global) {Global Quickest Change \\ Decision Maker};
    \node [right of=global,node distance=3.5cm] (globalaction)[draw=none]{};
    \node [above of=l2,left of =l2,node distance=1cm] (s2)[draw=none]{};
    \draw[->](global) -- node[above,pos=0.9]{$\{\text{change}, \text{continue}\}$}  (globalaction);
    \draw[->] (l2) -- (l3);      \draw[->] (l3) -- (l4); \draw[->](l4) -- (cdots);
    \draw [<->] (l2) --  (global);
    \draw [<->] (l3) --  (global);
        \draw [<->] (l4) --  (global);
      \end{tikzpicture}  

   \caption{\small  Quickest Change Detection Problem involving a single anticipatory agent acting repeatedly (or multiple anticipatory agents acting sequentially) and a global decision maker. 
    The anticipatory model for individual decision makers is discussed in Sec.\ref{sec:clmodel} and Sec.\ref{sec:structure} and results in time inconsistent decision making. The interaction of the  agents with a global decision maker to achieve  quickest detection  is detailed in  Sec.\ref{sec:qdnew}.}
  \label{fig:schematic}
\end{minipage}
}
\end{figure}
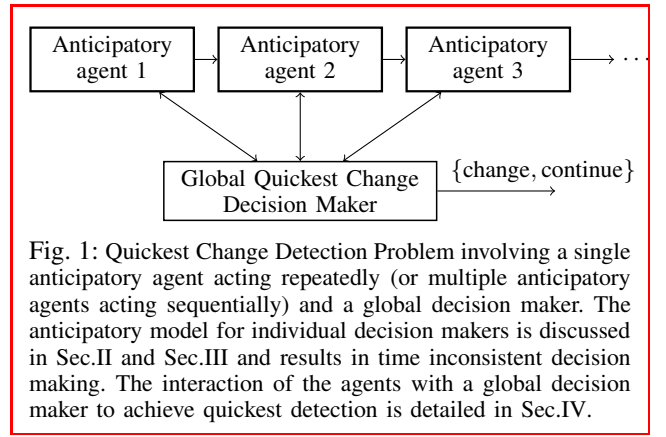

Figure \ref{fig:schematic} shows our schematic setup.
Anticipatory agents can mimic either strategic  human decision makers \cite{CL01} or an automated command-control decision system \cite{Ros12}.
The anticipatory agents  act sequentially and are affected by the decisions of previous agents.  A global decision maker monitors the decisions of these  anticipatory agents. How can the global decision maker use the local decisions from these anticipatory agents to decide when a change has occurred in the underlying state of nature?  The goal of the global decision maker is to achieve quickest change detection,  namely, minimize  the
Kolmogorov-Shiryaev criterion \cite{Shi63}, involving the   false alarm and decision delay penalty.

\subsection{Anticipatory Decision Making}

Anticipatory decision making has   applications in cyber-physical systems such as human-sensor, human-robot and command-control systems \cite{Ros12}.
Here are two  applications.\\
(i) {\em   Human decision makers}.
In behavioral economics,  Caplin \& Leahy
 \cite{CL01} propose a remarkable model for anticipatory human decision making
via a horizon-2  decision process: the first stage involves
choosing an action to minimize an anticipatory psychological  reward (involving the probabilities of choosing actions at stage 2),
while at the second stage the agent realizes its actual reward.
Such  anticipatory models  mimic   important features of human decision making:\\ \textbf{(i)} Extensive studies in   psychology, neuroscience \cite{BT16,CBS18}  show  that humans are  anticipation-driven, and even
simple decisions  involve sophisticated multi-stage planning.\\  {\bf (ii)}  Anticipatory agents act to reduce anxiety.   \cite{CB64}  presented  experimental results where people  chose a larger electric shock than waiting anxiously for a smaller shock.  \\ \textbf{(iii)} Anticipative agents  often deliberately  avoid information. \cite{MM83} reports that giving  patients more  information before a stressful medical procedure raised their anxiety.
 \\
(ii) {\em Level 3 Situation Awareness}. In defense   command-control systems, Level 3  Situation Awareness  (SA) \cite{End16}  involves the ability to project 
implications of future actions (plans).  Level~3 SA \cite{Bla13} is achieved through knowledge of  Levels 1 and 2 SA, and then extrapolating this information forward in time  (as an anticipatory reward involving probabilities of future actions) to determine how it will affect future decisions/plans~\cite{End16}. Prediction is concerned with guessing future states based on extensive training; in contrast, anticipatory decision making  \cite{KSP11} involves
preparing to respond to previously unseen scenarios. \cite{Lan86}  shows that many command and control systems overestimate their ability to react.

\subsection{Anticipatory Decision Making Yields  Time Inconsistency}
An important aspect of anticipatory decision making is \textit{time inconsistency}.
The dependence of the current
reward on future plans results in a deviation between planning and execution. This 
phenomenon is called  time-inconsistency\footnote{ In game-theoretic terms, time-inconsistency arises when the optimal
policy to the current multi-stage decision  problem is sub-game imperfect.} \cite{BM14} and 
 Bellman's principle of optimality no longer holds. 
Time inconsistency results in the 
{\em planning fallacy} of  Kahneman \& Tversky~\cite{KT79}: people tend to underestimate the time required to complete a future task.
Compared to rational agents, optimistic agents take higher risk of
making the wrong decision but have higher  anticipatory reward. 
\cite{BP05}
show that it is optimal for agents with anticipatory
reward to take irrational beliefs (referred to as subjective beliefs)
deliberately. This explains  the optimistic
planning fallacy, in which people tend to overestimate future rewards.
As will be  discussed below, the appropriate concept of optimality for  time-inconsistent problems  is the \textit{subgame Nash equilibrium}.

\subsection{Quickest Detection with  Anticipatory Agents}

Having motivated anticipatory decision making, we  turn to the second main idea of the paper, namely, \textit{Bayesian  quickest change detection by a global decision maker which uses
the  decisions of anticipatory agents} (local decision makers);  see Fig.\ref{fig:schematic}. In Bayesian quickest detection,
 the change time is specified by a prior \cite{PH08,TV05}.


We start by outlining important  applications that motivate the quickest detection problem with anticipatory agents.

The first class of examples involve social media based 
accommodation systems such as Airbnb.   Individuals with anticipatory feelings make   decisions whether to rent a  property; these decisions are affected by the reviews (decisions) of previous agents.
A  global decision maker (e.g. Airbnb) monitors these local decisions. {\em How can the global decision maker  detect if there is a sudden change in the demand for a specific  accommodation due to the presence of a new competitor?} The supplementary document  discusses
this example in detail. 

A related 
example  arises in the  measurement of the adoption of a new product using a micro-blogging platform like Twitter. 
The adoption of the technology diffuses through the market but its effects can only be observed through the tweets of select individuals of the population.
 These selected individuals  interact and learn from the decisions (tweeted sentiments) of  other members. Suppose the state of nature suddenly changes due to a sudden market shock or presence of a new competitor.
 The goal for a market analyst is to detect this change.
 
The  second  class of examples involves  anticipatory situation awareness  (SA) in a \textit{team} setting \cite{GCW06}.  For example,  \cite{EHF10}  introduced a situational
adapting system to assess team SA for fighter pilots
based
on information fusion. 
Suppose
individual SA systems monitor an enemy  target or enemy radar  (state). Given noisy measurements of the state,  each SA system (equipped with a Bayesian tracker) makes decisions about the threat  and  relays these  decisions to subsequent SA systems in the team. A global decision maker (supervisory system) monitors these decisions to assess overall  threat level.
How can the global decision maker  detect a sudden change in the threat?
Such a  change  is reflective of the enemy target making purposeful  maneuvers; or the enemy radar  switching modes between  search, acquisition or  track.
 

 The  third  example involves   human-sensor interface systems,
 where  anticipatory human decision makers are equipped with  sensing/computing devices. The sensing device  observes the state in noise. The computing device evaluates   the posterior distribution and provides the agent with these probabilities. The agent (human)  then makes  anticipatory decisions. The aim is to devise a change detection algorithm  that compensates for the anticipatory human decision maker.
 Such schemes are studied extensively in situation assessment of pilots 
\cite{HGW11} and validated based on 
simulations involving pilots performing a landing
approach  into an airport.
Other 
examples  include assistive care for the dementia \cite{BPH05}  where a machine monitors human decisions (activities) for changes in routine behavior indicating sudden onset of memory impairment.

 \subsection{Main Results}
 Sec.\ref{sec:clmodel}  reviews time inconsistent sequential decision problems and the framework for anticipatory decision making  as a 2-stage stochastic optimization problem. Due to the time inconsistency of the decision problem, the appropriate notion of optimality is the subgame Nash equilibrium policy. In  Sec.\ref{sec:structure}, our main contribution is to  introduce sufficient conditions on the anticipatory  model  so that the Nash equilibrium has a useful structure; see Theorem \ref{thm:nashstructure}.
This structure reveals several interesting features about anticipatory decision making. 

 Sec.\ref{sec:qdnew} formulates the quickest change detection protocol involving multiple anticipatory agents where a global decision maker uses the decisions of the anticipatory decision makers to decide if a state has changed.   The optimal policy that minimizes the Kolmogorov-Shiryaev criterion is formulated as the solution of a stochastic dynamic programming problem. Then
 Sec.\ref{sec:structure2} characterizes the structure of the Bayesian belief updates and achievable cost of the quickest detector without brute force computations. It derives  important  structural properties of the Bayesian updates of the local and global decision makers (Theorem \ref{thm:monotone} and Theorem \ref{thm:globalbelief}), constructs a lower bound for the optimal cost incurred using Blackwell dominance (Theorem \ref{thm:blackwell}), and presents numerical examples of the unusual structure of the optimal quickest change policy (non-convex stopping region) and non-concave value function.

\begin{figure}
\text{  \begin{subfigure}{.25\textwidth} 
 \resizebox{4.3cm}{!}{ \pgfplotsset{compat = 1.3}
  \begin{tikzpicture}  
\begin{axis}[width=6cm,height=3cm,ticks=none, y label style={at={(0.1,1.1)}},ylabel style={rotate=-90}
    ,xlabel=belief $\belief$
    ,ylabel=$\globalpolicy^*(\belief)$
    ,axis x line = bottom,axis y line = left
    ,ymax=0.8 
    ,ymin=-1.4
    ]
\addplot[blue,ultra thick] coordinates {(-1.1,0) (-0.2,0) (-0.2,-1) (1.1,-1)};
\end{axis}
  \draw[red,<->]    (1.8,1) -- (4.3,1) ;
\end{tikzpicture}}
\caption{Classical} \label{classical}
\end{subfigure}  \hspace{-0.6cm}
\begin{subfigure}{.25\textwidth}
\resizebox{4.3cm}{!}{ \pgfplotsset{compat = 1.3}
   \begin{tikzpicture}
\begin{axis}[width=6cm,height=3cm,ticks=none,
    ,xlabel= belief $\belief$,y label style={at={(0.1,1.1)}},ylabel style={rotate=-90}
    ,ylabel=$\globalpolicy^*(\belief)$
    ,axis x line = bottom,axis y line = left
    ,ymax=0.8 
    ,ymin=-1.4
    ]
    \addplot[blue,ultra thick] coordinates {(-1.1,-0.1) (-0.2,-0.1) (-0.2,-1) (0.4,-1)
      (0.4,-0.1) (0.7,-0.1) (0.7,-1) (1.1,-1)};
  \end{axis}
  \node[text width=3cm] at (0.8,1) {cont};
  \node[text width=3cm] at (0.8,0.2) {\red{stop}};
  \draw[red,<->]    (1.8,1) -- (3,1) ;
   \draw[red,<->]    (3.6,1) -- (4.5,1) ;
\end{tikzpicture}}
\caption{With anticipatory agents}
\label{qdjump}
\end{subfigure}}
\caption{Optimal Quickest Change Detection Policy $\globalpolicy^*$ as a function of  Bayesian belief  $\belief$. In classical quickest detection, the stopping set is convex (connected). In comparison, for quickest detection with anticipatory agents (this paper), the stopping set is nonconvex (disconnected) as indicated in red.}
\label{fig:qdintro}
\end{figure}
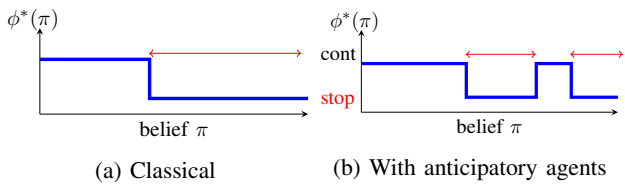

In classical quickest detection
 \cite{Shi63,PH08}, the optimal  policy has a  threshold structure: when the posterior
probability of change exceeds a threshold, it is optimal to declare a change; see 
Fig.\ref{classical}. The  stopping
set (set of posteriors probabilities where it is optimal to declare “change”) is convex.
\blue{In quickest detection involving a global decision maker interacting with anticipatory  agents (this paper)}, the remarkable feature is that the stopping set is disconnected, see Fig.\ref{qdjump}.  One sees  the counter-intuitive property: the optimal detection policy switches from
announce “change” to announce “no change” as the posterior probability  of a change increases!
Thus making a global decision as to whether a change has occurred based on local decisions of interacting
agents is non-trivial.

\subsection{Perspective on Main Results} To give additional perspective on the main results discussed above, we now briefly discuss  important insights regarding anticipative decision  makers in   a quickest detection framework.

{\em 1. Social Learning.}   
The anticipatory model used in this paper is  from 
\cite{CL01}; see also \cite{BP05,BPP17}. This  
generalizes classical social learning models that have been studied extensively in sociology, economics and  signal processing  \cite{Cha04,Kri12,KP14,BMS20}. Classical social learning  assumes that agents make one-shot (myopic) decisions to maximize their expected utility. The behavioral economics models  considered here are useful  generalizations of social learning  since they involve multi-stage planning; as mentioned earlier,  even simple human decisions  involve multi-stage planning with time-inconsistency.

Our sequential framework of  multiple  decision makers is similar  to 
team decision theory  \cite{Tsi93,VV97}; the key difference being  time inconsistency.

\blue{This paper  differs from \cite{Kri12} where quickest detection was considered with myopic social learning based local decisions. Motivated by behavioral economics \cite{CL01},  we   consider
  a 2-stage decision framework for each local decision maker that is  more general than myopic social learning. This 2-stage framework captures several salient features of human decision-making including anticipation, time inconsistency and deliberate avoidance of information.
   Also, in our quickest detection formulation, the jump change  affects both the rewards of the agents and the transition kernel of the physical state (in the myopic case \cite{Kri12},  there is no transition mechanism).
   Our constructed model  ensures that
   we can seamlessly use the behavioral anticipatory model of \cite{CL01} without modification.}


{\em 2. How un-informed local decision makers affect global   decision making?}
In order to optimize its change detection policy, the global decision maker must interpret decisions of the  local decision makers, knowing that the local decision makers  are  anticipatory and that they use  decisions from previous agents.  A well known characteristic of this sequential multi-agent framework is that agents herd \cite{Cha04} - they ignore their own observations and parrot decisions of previous agents.   The multi-threshold structure of the global decision maker's  optimal   policy (Figure \ref{qdjump}) can be interpreted as saying that  the global decision maker  acts in a non-trivial  manner to compensate for the poorly informed local  decision makers. In comparison, the classical threshold policy (Figure \ref{classical}) results when the local decision makers are well informed (exchange their posterior distributions rather than anticipatory actions); see \cite{Kri12} for discussion in terms of  Bayesian social learning.


{\em 3. Change Blindness.} 
The multi-threshold change detection  policy in Fig.\ref{qdjump} can be interpreted as a form of {\em change blindness}, namely, people fail to detect surprisingly
large changes to scenes \cite{SR05}. Even though the posterior probability of a change is higher than a change threshold, the optimal behavior indicated is to detect no change.

{\em 4.  Deliberate Avoidance of Information.}  Theorem \ref{thm:nashstructure} in Sec.\ref{sec:structure} shows that the subgame Nash equilibrium at time 1 has a bang-bang structure.  It justifies the observation \cite{CL01}  that agents with anticipatory emotions may choose to deliberately  avoid information. As mentioned earlier,  \cite{MM83} reports that giving some patients more  information before a stressful medical procedure raised their anxiety. \cite{CBS18} shows that 
humans 
selectively treat the opportunity to gain knowledge
about future favorable outcomes, but not unfavorable outcomes.

Finally, we  emphasize that humans  likely do not solve 
time inconsistent  decision processes to make decisions.
The  time inconsistent
behavioral economics models in  \cite{CL01,BP05,BPP17} are widely used  because  they provide {\em generative} models for   the peculiarities of   anticipatory human decision making.

\vspace{-0.42cm}

\subsection{Organization}
 The paper is organized  into three  inter-related parts:
\begin{compactenum} \item {\bf Part~1} deals with anticipatory  models for a single decision maker and characterizes the Nash equilibrium.
\item {\bf Part 2} of the paper deals with  quickest  change detection with a team of anticipatory decision makers.
  \item {\bf Supplementary Material} (separate submitted document) contains proofs of theorems and a detailed tutorial example of anticipatory decision making in  social media.
  \end{compactenum}
\hrule

\vspace{0.5cm}

{\large {\bf Part 1. Anticipatory Models and  Nash Equilibrium}} 

Sec.\ref{sec:clmodel} formulates  anticipatory decision making. Sec.\ref{sec:structure} characterizes the structure of the Nash equilibrium with examples.

\section{Anticipatory Decision Making}
\label{sec:clmodel}
This section defines time inconsistent decision problems and reviews the influential behavioral economics model \cite{CL01}  for human decision making with anticipatory feelings. This model will be used in
Sec.\ref{sec:qdnew} to formulate our human sensor interactive quickest change detection problem.

\subsection{Time Inconsistent Sequential Decision Problems}
\label{sec:inconsistent}
We start with a brief discussion of time inconsistent decision problems; see \cite{BM14} for an exposition.
Let $\{\generalstate_k,k= 1,\ldots,\horizon\}$ denote a controlled Markov chain evolving on a finite time horizon size $\horizon$. The initial distribution for $\generalstate_1$ is denoted as  $\belief_1$.  Let  $\pol_k$ denote a (possibly randomized) decision policy that maps the state $\generalstate_k$ to
an action $\action_k$ at time $k$.
For $n=1,2,\ldots,\horizon$, define the expected  utility-to-go
\beq  \utilitytogo_n(\generalstate_n,\vpol_{n:\horizon}) = \E_{\vpol_{n:\horizon}} \bigl\{ \sum_{k=n}^{\horizon}  \reward_{n,k}\big(\generalstate_n,\generalstate_k,\pol_k(\generalstate_k)  \big) \bigr\}
\label{eq:bjork}
\eeq
The aim is to compute the policy sequence $\argmax_\vpol \utilitytogo_n(\generalstate_n,\vpol_{n:\horizon})$.
As  the reward $\reward_{n,k}$ depends  on $n$ and $k$,  and also $\generalstate_n,\generalstate_k$, this optimization  problem is  {\em time inconsistent} since  the principle of optimality (Bellman's dynamic programming equation) does not hold; see~\cite{BM14}.

\subsubsection{Subgame Perfect Nash Equilibria}

As discussed in \cite{BM14}, an  appropriate
method of ``solving'' a
time inconsistent problem is in 
game-theoretic terms.\footnote{The following intuitive argument from  \cite{BM14} is helpful: Looking to maximize $\utilitytogo_n(\generalstate,\vpol_{n:\horizon})$ over the class of policies restricted to $[n:\horizon]$, a player at time $n$ would like in principle to maximize $\utilitytogo_n(\generalstate,\vpol_{n:\horizon})$ over $\pol_n,\ldots,\pol_{\horizon}$.
But the player at time $n$ can only choose the policy $\pol_n$ - so the maximization is not possible. Instead of looking for optimal feedback laws, in a time inconsistent problem one considers the subgame perfect Nash equilibrium.}
\begin{compactenum}
\item
  Given state $\generalstate_{\horizon} = \generalstate$, player $\horizon$ chooses policy
  \beq \pol^*_{\horizon}(\generalstate) = \argmax_{\action_{\horizon}} \utilitytogo_{\horizon}(\generalstate,\action_{\horizon})
\label{eq:dp_modified_last}
\eeq
This yields the value function $\valueb_{\horizon} =\utilitytogo_{\horizon} (\generalstate,\pol^*_{\horizon}).$
\item Given  $\generalstate_{\horizon-1}=\generalstate$, and that player $\horizon$
  is using policy $\pol^*_{\horizon}$, player $\horizon-1$ chooses policy
  \begin{align}  \pol^*_{\horizon-1}(\generalstate) &= \argmax_{\action_{\horizon-1}} \utilitytogo_{\horizon-1}(\generalstate,  \action_{\horizon-1},\pol^*_{\horizon}) \nonumber
  \\  \valueb_{\horizon-1} (\generalstate) &= \utilitytogo_{\horizon-1}(\generalstate,\pol^*_{\horizon-1},\pol^*_{\horizon})
                                               \label{eq:dp_modified_step}
  \end{align}
    \item Proceed by backward induction to compute policies $\pol^*_{N-2},\ldots,\pol^*_{1}$.
  \end{compactenum}
  The above procedure is called the {\em extended} Bellman equation in \cite{BM14}.
  The  sequence of policies $\vpol^*=(\pol^*_1,\ldots,\pol_{\horizon}^*)$, constitutes a subgame perfect Nash equilibrium; see \cite{BM14} for details.

 \subsubsection{Remarks}  (i) As might be expected, for  the time consistent case where $\reward_{n,k}(\generalstate_n,\generalstate_k,\action_k) = \reward_k(\generalstate_k,\action_k)$ in (\ref{eq:bjork}), the extended Bellman's equation becomes the standard Bellman's dynamic programming equation. \\ (ii) For the time inconsistent case, neither the Nash equilibrium  $\vpol^*$ nor its value 
 $\utilitytogo_n(\vpol^*)$ are unique. This is in contrast to time consistent dynamic programming  where the optimal policy may not be unique but the optimal value is always unique.
   
\subsection{Anticipatory Model of Caplin \& Leahy \cite{CL01}} \label{subsec:clmodel}
We now review  the time inconsistent model  for anticipatory human decision making in  Caplin \& Leahy's  paper~\cite{CL01}. Their model uses the terminology of temporal lotteries in dynamic choice theory
\cite{KP78}.
We  translate their model  to a more familiar Markov decision process framework. While the messy notation below is unavoidable, the reader should keep in mind that the final outcome is a time inconsistent problem of the form (\ref{eq:bjork}) with horizon $\horizon=2$.
A  key step in the formulation  below is the anticipatory  state (\ref{eq:psych}) at time 1 which depends on the probability of future actions (at time 2); this gives the model its anticipatory property.

\subsubsection{Anticipatory Model and Time Inconsistency} The anticipatory decision model in \cite{CL01}
comprises two time steps indexed by $k=1,2$.
The physical state 
$\physical_k \in \physicalstatespace$, $k=1,2$, where $\physicalstatespace$ denotes the state space, evolves with Markov transition kernel $\pdf(\physical_2|\physical_1)$. Let $\action_1\in \actionspace_1$ and $\action_2\in \actionspace_2$ denote the actions  taken by the agent (human)
at time 1 and 2. These actions are determined by the non-randomized policies $\pol_1$ and $\pol_2$ where
\begin{equation} \label{eq:pol12}
\begin{split}
\action_1 &= \pol_1(\physical_1), \quad
\action_2 = \pol_2(\physical_2,\action_1).
\end{split}
\end{equation}

The first key idea in
Caplin \& Leahy
  \cite{CL01} is to define the anticipatory (psychological) state  $\psych_k$,  $k=1,2$:
  \begin{equation} \begin{split}
  \psych_1 &= \fun\big(\physical_1,\action_1,\{\pdf(\action_2=\action|\physical_1,\action_1,\policy_2), \action \in \actionspace_2\}\big), \\
\psych_2 &= (\physical_2,\action_2,\action_1), 
\end{split}
\label{eq:psych}
\end{equation}
for some pre-defined function $\fun$. \blue{Note  $\policy_2$ is a deterministic function that parametrizes $\pdf(\action_2=\action|\physical_1,\action_1,\policy_2)$.}
In \cite{CL01}, $\psych_k$   models the human decision maker's state of mind (anxiety). More generally,
$\psych_k$ can model any anticipatory plan, such as for example in situation awareness systems.
Note that the anticipatory state $\psych_1$ depends on 
the set of conditional probabilities    $\{\pdf(\action_2=\action|\physical_1,\action_1,\pol_2), \action\in \actionspace_2 \}$. These conditional probabilities  model anticipation (anxiety)\footnote{As discussed in \cite{CL01},  introducing anticipatory emotions explains why changing an outcome from zero to a small positive number can have a large effect on anticipation. Human decision makers are sensitive to the possibility rather than probability of negative outcomes \cite{LWH01}.
 A  terrorist attack (unlikely event) worries people a lot more than a car crash (high probability event).}
of the decision maker at time~1 about possible actions it can make at time 2.
  The anticipation is resolved  at time~2 when physical state $\physical_2$ is observed and all uncertainty is resolved; hence  the anticipatory  state $\psych_2$ only  contains  physical state $\physical_2$ and  realized action $\action_2$.

The next  key idea in \cite{CL01} is that the anticipatory agent makes  decisions by
 maximizing  the 2-stage  anticipatory   utility 
 \begin{align}
   \sup_{\pol_1,\pol_2}  \utilitytogo({\pol_1,\pol_2}) &= \E_{\pol_1,\pol_2}\{\reward_1(\psych_1) + 
                                              \reward_2(\psych_2) \}  \label{eq:cl_cost}
 \end{align}
 Here $\reward_k(\psych_k) \in \reals$ denote the reward functions.
 The 2-stage anticipatory utility, called  psychological  utility in \cite{CL01},  (\ref{eq:cl_cost})  looks just like a standard time separable utility except for the presence of the anxiety term $\{\pdf(\action_2=\action|\physical_1,\action_1,\pol_2),\action\in \actionspace_2\}$  in $\reward_1(\psych_1)$.
 This $\pol_2$ dependency  gives rise to time inconsistency in  decision making. Indeed (\ref{eq:cl_cost})  is a special case of the general time inconsistent formulation
 (\ref{eq:bjork}) with
\begin{align} \label{eq:rewards12}
 \reward_{2,2} &= \reward_2(\physical_2,\action_2,\action_1), \; \reward_{1,1}=0, \\ \reward_{1,2} &=
\reward_1\big(\fun(\physical_1,\action_1,\{\pdf(\action_2=\action|\physical_1,\action_1,\pol_2),\action\in\actionspace_2\})\big) + \reward_{2,2}\nonumber
\end{align}
\blue{As in \cite{CL01}, we assume that the agent knows all the parameters in the above anticipatory model. The key point is that  the reward at time 1 depends on the psychological (anticipatory) state which in turn depends on the probability of future actions and states.}

\subsubsection{Subgame Perfect Nash Equilibrium} Caplin \& Leahy \cite{CL01} `solve' the  time inconsistent decision problem (\ref{eq:cl_cost}) using the extended Bellman equation described in Sec.\ref{sec:inconsistent}. Indeed, 
the optimal policy at time 2 simply follows from (\ref{eq:dp_modified_last}) with $\horizon = 2$:
\beq  \pol_2^*(\physical_2,\action_1)  
=\argmax_{\action_2}  \reward_2(\physical_2,\action_2,\action_1) 
\label{eq:period2} \eeq
Note that by definition (\ref{eq:pol12}),  $\pol_2^*$ depends on $\action_1$ and $\physical_2$.

To specify the optimal policy at time 1, we first introduce the following compact notation. Define 
\begin{equation} \begin{split}
\cond_\action &\ole \int_\physicalstatespace I(\physical_2: \pol_2^*(\physical_2,\action_1) = \action)  \, \pdf(\physical_2|\physical_1)\, d\physical_2, \\
  \cond &= \{\cond_\action,  \action \in \actionspace_2\} 
\end{split} \label{eq:cond}
\end{equation}
At time 1, due to time inconsistency,  the agent chooses a time consistent policy $\pol_1^*$ based on
 extended Bellman equation (\ref{eq:dp_modified_step}):
\begin{align}  
  \pol_1^* (\physical_1)&= \argmax_{\action_1}  \utilitytogo_1(\physical_1,\action_1,\pol_2^*), \label{eq:period1}  \\
\valueb_1(\physical_1) &= \max_{\action_1}  \utilitytogo_1(\physical_1,\action_1,\pol_2^*), \nn
  \\
\utilitytogo_1(\physical_1,\action_1,\pol_2^*) &=   \blue{\reward_1\big(\fun(\physical_1,\action_1,
\cond )
\big)  + \E\{\reward_2(\physical_2,\action_2,\action_1)| \physical_1, \action_1,\ \pol_2^*\}} \nn \\
& \hspace{-1.5cm} = \reward_1\big(\fun(\physical_1,\action_1,
\cond ) \big)+ 
  \int_\physicalstatespace  \reward_2\big(\physical_2,\pol_2^*(\physical_2,\action_1) ,\action_1\big) \, \pdf(\physical_2| \physical_1)\, d \physical_2   \nn
\end{align}
Recall $\pdf(\physical_2|\physical_1)$ is the transition kernel of the physical state.

\noindent {\bf Remarks}: 
(i)  (\ref{eq:period1}) is identical to the  master equation  \cite[Eq.2]{CL01}. Indeed,  in more  compact notation we can write (\ref{eq:period1}) as
\begin{equation}
  \pol_1^* (\physical_1)= \argmax_{\action_1} \{ \reward_1\big(\fun(\physical_1,\action_1, \cond)\big)  + 
  \E_\cond \{  \reward_2\big(\physical_2,\action_2,\action_1) \}
\label{eq:cleqn}
\end{equation}
which is the same as the  master equation  \cite[Eq.2]{CL01}   since 
\begin{multline*}  \E_\cond \{  \reward_2\big(\physical_2,\action_2,\action_1) \}   =
  \int_{\actionspace_2} \int_\physicalstatespace \reward_2(\physical_2,\action,\action_1) \cond_\action \, d\physical_2 d\action \\ = 
   \int_{\actionspace_2}\int_\physicalstatespace  \reward_2\big(\physical_2,\action,\action_1)\, I(\big(\action = \pol_2^*(\physical_2,\action_1) \big)\, \pdf(\physical_2| \physical_1) d \physical_2 d\action \end{multline*}
 (ii) The  anticipatory (psychological) state $\psych_1$ in
 (\ref{eq:psych}) consisted of  the set of conditional probabilities $\{\pdf(\action_2=\action|\physical_1,\action_1,\policy_2), \action \in \actionspace_2\}$. More generally,  one can formulate the anticipatory  state  with these conditional probabilities replaced by
 \beq \{ \E\{\ant(\action_2=\action,\physical_2) | \physical_1,\action_1,\pol_2\}, \action \in \actionspace_2\}
 \label{eq:ant}\eeq
 for some pre-defined function $\ant$. As an example (which is  elaborated on in the supplementary material)
 $$ \psych_1 = \max\{ \pdf(\action_2=1|\physical_1,\action_1,\pol_2), \E\{\physical_2 I(\action_2 = 2) | \physical_1,\action_1, \pol_2)\} $$

\noindent (iii) 
We mentioned previously  that  the subgame Nash equilibrium approach to time inconsistency disregards
the fact that $\pol_2^*$ is no longer optimal at time 1. Another insightful way of viewing this is that
 the   estimated  anticipatory
reward $\reward_1\big(\fun(\physical_1,\action_1, \cond)\big)$ requires the agent to extrapolate what might happen at the second
stage, plans are not optimal once an action is taken. As an example,  people tend to assign higher future workload than what they
will actually take on.


{\em Summary}.  The key point in anticipatory decision making is the presence of probabilities of choosing future actions in the current reward, as depicted in the anticipatory  state~(\ref{eq:psych}). As a result,
maximizing the 2-stage anticipatory  utility  (\ref{eq:cl_cost}) is a time inconsistent problem. 
The anticipatory decision maker chooses actions $\action_1, \action_2$  according to policies $\pol_1^*$ in (\ref{eq:period1}) and
$\pol_2^*$ in  (\ref{eq:period2}); these policies constitute a subgame perfect Nash equilibrium. Indeed
 (\ref{eq:cleqn}) corresponds to the key master  equation (2) in \cite{CL01}.
The paper \cite{CL01} has received significant attention in  behavioral economics (mindful economics \cite{BT16}),  
neuroscience and psychology \cite{CBS18}.

  \subsection{Example 1. Financial Investment and Anticipatory Betting}
  \label{sec:bet}
The following example (based on \cite{CL01})  presents  anticipatory decision making in a simplified setting to illustrate rapidly  the key ideas. The problem is time inconsistent since  the utility at time 1 depends on the  expected physical state at time~2.

There are two periods.
An investor makes two decisions denoted $\acta$ and $\actb$
in period 1 (this simplifies the problem). 
\begin{compactenum} \item The decision
 $\acta \in \{\tv,\garden\}$ is 
whether to invest in short term stock  or long term bonds. If $\acta = \tv$, then the agent chooses
 $\actb \in [0,\wealthmax]$, namely how many  units to invest  in $\tv$, where
$\wealthmax$ denotes the initial wealth.
\item The physical state  $\physical_1$ denotes the probability
that $\tv$ yields a  return $\win$. For simplicity, assume $\physical_1 = 1/2$.

\item At time 2, 
the physical state $\physical_2 \in \{\win, \lose\}$ denotes whether the return on $\tv$ is satisfactory or not.

\item If the investor chooses $\acta = \tv$, invests $\actb$, and
the return $\physical_2$ is $\win$, then it earns $2 \actb$; so its 
wealth at the end of period 2 is $\wealth=\wealthmax + \actb $. If the return
is $\lose$, the investor loses $\acta$ and its wealth at the end of period 2
is $\wealth=\wealthmax - \actb$.

\item If the investor chooses $\acta = \garden$, then it invests the entire
$\wealthmax$ and obtains a return of $\wealthmax + \gardenreward$, where
$\gardenreward$ denotes the interest payment.
\end{compactenum}
Given final wealth $\wealth$, assume the agent's utility at time~2 is
\begin{equation} \label{rau2}
      \reward_2(\physical_2,\actb,\acta) =  \wealth - \beta\, \wealth^2 
    \end{equation}
    This utility models a risk averse agent with quadratic penalty loss  (which is used widely in behavioral economics).
    
We assume that the  agent's anticipatory utility at time 1 is
\begin{align}
    \utilitytogo_1(\physical_1,\acta=\tv,\actb) &=  \alpha(u_\Ali + \actb - \beta \actb^2) + \E\{  \reward_2(\physical_2,\actb,\acta)| \physical_1\}\nonumber \\
    \utilitytogo_1(\physical_1,\acta=\garden,\actb)&= g + \E\{  \reward_2(\physical_2,\actb,\acta)| \physical_1\}  \label{eq:Jex1}
\end{align}
where $u_\Ali,\alpha, g ,\beta$ are positive constants.
This decision problem is time inconsistent since the utility  at time 1 depends on the expected  physical state $\physical_2$.
Recall that  decisions $\acta,\actb$ are made at time 1 only (so there is no $\pol_2^*$ in \eqref{eq:period1}).
The  term $\alpha(u_\Ali + \actb - \beta \actb^2)$ is the excitement (suspense) of investing~$\actb$; the
    term $-\beta \actb^2$ models  the risk averseness of the agent.

    Let us work out $ \utilitytogo_1$ in (\ref{eq:Jex1}) explicitly.
Since the probability of $\tv$ returning $\win$ is 1/2, clearly
\begin{equation} \begin{split}
   \E\{ \reward_2(\physical_2,\actb,\acta=\tv) | \physical_1 \} &= 
   \wealthmax - \beta (\wealthmax^2 + \actb^2) \\
    \E\{ \reward_2(\physical_2,\actb,\acta=\garden) | \physical_1 \} &= 
    \wealthmax+\gardenreward - \beta\, (\wealthmax+\gardenreward)^2 
  \end{split} \label{eq:r2s}
\end{equation}
Therefore the optimal investment $\actb$ is zero if only the second period expected utility is considered.
The utility $\utilitytogo_1$
in \eqref{eq:Jex1} captures the tradeoff between 
the excitement and future anticipatory gain/loss,  leading to a time inconsistent problem.

The time consistent optimal policy  at time 1 using \eqref{eq:period1} is:
\begin{align}
    &\pol_1^* (\physical_1) = (\acta^*,\actb^*) \label{eq:polbet} \\
    & \actb^* = \argmax_{\actb\geq 0} \alpha u_\Ali + \alpha \actb - (1+\alpha) \beta \actb^2 = \frac{\alpha}{2(1+\alpha) \beta} \nn \\
    &\acta^* = \begin{cases} \tv &\text{ if } \gardenreward(1 - 2\beta\wealthmax) - \beta
      \gardenreward^2+g < \alpha(u_\Ali + \frac{\alpha}{4(1+\alpha) \beta}) \\
      \garden & \text{ otherwise } 
    \end{cases} \nn
\end{align}


\subsection*{Anticipatory Betting/Gambling \cite{CL01}}

We now describe an example involving  anticipatory betting/gambling
\cite{CL01}.
The setup is a special case of above. An agent chooses action
$\acta \in \{\bet,\nobet\}$.  $\actb \in [0,\wealthmax]$ denotes how much money is bet. The physical state $\physical_1 = P(\winbet)=1/2$, namely,  anticipated probability of win at stage 1, and  $\physical_2 \in \{\winbet,\losebet\}$ denotes the actual outcome at stage 2.
\begin{compactenum} \item
If the agent chooses $\acta=\bet$ then the final wealth is $\wealthmax+\actb$  if the bet is won ($\physical_2 = \winbet$) or $\wealthmax - \actb$ if the bet is lost ($\physical_2 = \losebet$). 
\item If the agent chooses $\acta=\nobet$, then the final wealth remains $\wealthmax$
  (instead of $\wealthmax+ \gardenreward$, i.e., interest $\gardenreward = 0$).
\item The risk averse utility at stages 2 and 1 are \eqref{rau2}, \eqref{eq:Jex1}
 with $\gardenreward=0$,
  $\bet$ replacing $\tv$ and $\nobet$ replacing $\garden$.
  \end{compactenum}
  Then \eqref{eq:r2s} holds with $\gardenreward=0$.
  The Nash equilibrium policy
  $\pol_1^* (\physical_1)$ is  (\ref{eq:polbet})  with $\gardenreward=0$,
  $\bet$ replacing $\tv$,  $\nobet$ replacing $\garden$.
  Sec.\ref{sec:spot} illustrates this model in  quickest detection.

\subsection*{Implications of Anticipatory Investment/Betting} The  agent  chooses $\tv$ (or $\bet$) even though it loses in terms of the risk averse final utility \eqref{eq:r2s},
yet individuals gamble because it heightens suspense (anticipation) prior to the resolution of uncertainty in the second stage. This illustrates  the time inconsistency of the problem: in the final period it is not useful to invest in $\tv$  (or $\bet$). Yet the investment is made at stage 1 with anticipatory feelings rather than the ultimate outcome;
 see \cite{CL01} for implications in  gambling/betting. To quote Samuelson \cite{Sam52}:  ``I am satisfied that a large fraction of the sociology of gambling and of risk taking will never significantly be discernible in
terms of the money prizes alone, as distinct from elements of
suspense....”

\section{Characterizing the Nash Equilibrium Policy of Anticipatory Decision Maker and Examples} \label{sec:structure}

The previous section gave a general setup of the anticipatory decision making model and associated subgame Nash equilibrium policy.  However,  the Nash equilibrium (\ref{eq:cleqn}) is the solution of the extended Bellman equation (integral equation) and is difficult to compute in general.  In this section, our main contribution is to
 make specific assumptions on the anticipatory model
to  give a  useful  characterization of the  Nash equilibrium. Specifically,
these assumptions result in a bang-bang and threshold structure for  the  subgame Nash equilibrium  policy
(Theorem \ref{thm:nashstructure} below).
\blue{This structural result illustrates the optimality of simple decision-making rules and will be illustrated by an example involving  situation awareness.}

\subsection*{Bayesian parametrization of transition kernel and reward}
Recall $\reward_2$ is the reward at time 2; see  (\ref{eq:psych}),  (\ref{eq:cl_cost}). In the rest of the paper, we will parametrize  $\reward_2$ and the transition kernel $\pdf(\physical_2|\physical_1)$ by a Bayesian parameter. 
The parameterized reward and transition kernel are constructed as follows:
Define the  reward $\reward_2(\physical_2,\action_2,\action_1,\state)$  and transition kernel $\pdf(\physical_2|\physical_1,\state)$ which now also depends on a state of nature  (ground truth) $\state$. The process $\state \in \statespace = \{1,2,\ldots, \beliefdim\}$ will be formally defined  in Sec.\ref{sec:qdnew}
to model  change in quickest detection. Then define the parametrized reward $\rewardp$ and transition kernel
$\pdfp(\physical_2|\physical_1)$ 
as
\beq
\begin{split}
\rewardp(\physical_2,\action_2,\action_1) &= \sum_{\state \in \statespace} \reward_2(\physical_2,\action_2,\action_1,\state)\, \private(\state) \\
\pdfp(\physical_2|\physical_1) &= \sum_{\state \in \statespace} \pdf(\physical_2|\physical_1,\state)\, \private(\state)
\end{split}
\label{eq:rewardparam}
\eeq
Here $\thbelief$ is an $\beliefdim$-dimensional  Bayesian belief (posterior) vector that  lies in the unit $\beliefdim-1$ dimensional simplex~$\Belief$ of probability mass functions:
$\thbelief = [\thbelief(1),\ldots\thbelief(\beliefdim)]^\p \in \Belief $, where
\begin{equation} \begin{split}
\Belief &= \{\thbelief:  \thbelief(i) \in [0,1], \quad \sum_{i=1}^m \thbelief(i) = 1 \}
\end{split}
\label{eq:thbelief}
\end{equation}
\blue{The  posterior $\thbelief$  will be formally defined in (\ref{eq:public})  and appears naturally  in   the quickest change  detection formulation in Sec.\ref{sec:qdnew} (where the underlying state pf nature  $\state$ jump changes). In this section, $\thbelief$ is simply a fixed probability vector in the two-stage anticipatory decision model discussed above.}

\subsection{Structural Characterization  of Nash equilibrium}

With $\rewardp$ defined in (\ref{eq:rewardparam}),
for notational convenience, define
  \beq \diffreward(\physical_2,\action_1) = \rewardp(\physical_2,2,\action_1) - \rewardp(\physical_2,1,\action_1)
  \label{eq:diffreward}
  \eeq

We make the following assumptions on the anticipatory decision model of  Sec.\ref{subsec:clmodel}:

\begin{enumerate}[label=(A{\arabic*})]
  

\item \label{actionspace}  The action spaces are $\actionspace_1 = [0,1]$, $\actionspace_2= \{1,2\}$. Recall the actions
  $\action_1 \in \actionspace_1$ and $\action_2 \in \actionspace_2$.\\
  The state space is  $ \physicalstatespace = [0,1]$. Recall $\physical_1,\physical_2 \in \physicalstatespace$.
  \item \label{convexr2} $\rewardp(\physical_2,\action_2,\action_1)$ is convex in $\action_1$.
\item \label{supermod} 
   $\diffreward(\physical_2,\action_1)   $ defined in (\ref{eq:diffreward})  is  increasing in $\physical_2$. Equivalently, 
   $\rewardp(\physical_2,\action_2,\action_1)$ is supermodular in $(\physical_2,\action_2)$.
 \item \label{implicit}
   The solution $\physical_2^*(\action_1)$ of $\diffreward(\physical_2,\action_1)= 0$ exists for $\action_1 \in (0,1)$ and  is  continuously differentiable  on $(0,1)$.
\item   \label{convexthreshold}     $ \frac{\partial \diffreward}{\partial \action_1} \frac{\partial^2 \diffreward}{\partial \physical_2 \partial \action_1} - \frac{\partial \diffreward}{\partial \physical_2} \frac{\partial^2\diffreward}{\partial \action_1^2}\geq 0$
  \item \label{beta} The anticipatory reward is $\reward_1(\psych_1) = \beta \psych_1$ where  $\beta>0$ and the 
  psychological state (see (\ref{eq:ant})) is
$$ \psych_1 = \max\{ \E\{\ant(\action_2=\action,\physical_2) | \physical_1,\action_1,\pol_2\}, \action \in \actionspace_2\}$$
\item \label{convexcdf} 
$ \ant(\action_2=1,\physical_2)\, \pdfp(\physical_2|\physical_1) $ is increasing in $\physical_2$ \\
$\ant(\action_2=2,\physical_2) \,\pdfp(\physical_2|\physical_1)$ is decreasing in $\physical_2$.
\end{enumerate}

\vspace{0.5cm}

The following  structural result  characterizes the structure of the subgame Nash equilibrium. For subsequent reference, we will denote the explicit dependence of  $\pol_1^*$ and $\pol_2^*$ on Bayesian parameter $\thbelief$ (see (\ref{eq:thbelief})) as $\polp1$ and $\polp2$.

\begin{theorem} \label{thm:nashstructure}
  Consider the anticipatory decision model of  Sec.\ref{subsec:clmodel} with action and state spaces specified by
\ref{actionspace}.  Then
\begin{compactenum} \item
Under   \ref{supermod},  \ref{implicit}, the subgame perfect Nash equilibrium policy $\pol_2^*$ specified by (\ref{eq:period2}) has a  threshold structure:
\begin{equation} \label{eq:pol2threshold}
    \polp2(\physical_2,\action_1) = \begin{cases} 1 & \text{ if } \physical_2 \leq \physical_{2,\thbelief}^*(\action_1) \\
      2 & \physical_2 > \physical_{2,\thbelief}^*(\action_1) 
    \end{cases}
  \end{equation}
  for some threshold state $\physical_{2,\thbelief}^*(\action_1) \in [0,1]$ which depends on the Bayesian parameter $\thbelief$.
  \item 
  Under \ref{implicit}, \ref{convexthreshold},  threshold state  $\physical_{2,\thbelief}^*(\action_1)$ is convex in~$\action_1$.
\item
Under \ref{convexr2}-\ref{convexcdf},
the utility-to-go $\utilitytogo_1(\physical,\action_1,\pol^*_2)$ defined in (\ref{eq:period1}) is convex in $\action_1$.
  Therefore, the subgame Nash equilibrium policy $\pol^*_1$ has the following bang-bang\footnote{The phrase ``bang-bang controller'' comes from classical optimal control theory. It characterizes a control policy with continuous-valued actions that switches between two extremes.} structure:
  \begin{equation}
  \polp1(\physical_1) = \begin{cases} 1 & \text{ if }  \beta > \beta^* \\
      0  & \text{otherwise} 
    \end{cases} \label{eq:bangbang}
  \end{equation}
  for some positive constant $\beta^*$.
  ($\beta$ is defined in \ref{beta}.)
\end{compactenum}
\end{theorem}

The proof is in the supplementary document.

\blue{{\em Deliberate Avoidance of Information.} The structure of the Nash policy in Theorem \ref{thm:nashstructure} yields  interesting consequences. Suppose $\action_1$ denotes a non-refundable financial deposit made by the agent at time 1 in anticipation of  choosing action $\action_2=1$ at time~2. Due to the bang-bang structure of (\ref{eq:bangbang}) the agent makes a full deposit $\action_1 = 1$ if $\beta > \beta^*$.  Yet this full non-refundable deposit does not guarantee that the agent will choose $\action_2 = 1$ since if $\physical_2 > 
 \physical_2^*(\action_1) $, then  the agent will choose $\action_2 =2 $. Thus the agent would like to avoid observing $\physical_2$. There is an elegant interpretation of this in \cite{CL01}, namely, the agent might deliberately choose not to observe the state $\physical_2$ in order not to lose the deposit.  ``In this manner, anticipatory emotions may rationalize the deliberate avoidance of information'' \cite{CL01}.}

\subsection{Discussion of Assumptions}
\label{sec:discussion_of_assumptions}
   Assumptions \ref{actionspace}-\ref{convexcdf} are generalizations of (and therefore weaker than)  the assumptions in  \cite{CL01}, where an example of  anticipatory decision making for choosing a holiday destination is discussed.  Note that \ref{convexr2} to \ref{convexthreshold} are assumptions on $\rewardp$, while \ref{beta},\ref{convexcdf} are assumptions on $\reward_1$.
\\
{\bf \ref{actionspace}}:
In \cite{CL01} and also the social media accommodation example (supplementary material), $\actionspace_1= [0,1]$ denotes the feasible set of deposits made to secure an accommodation, while $\actionspace_2=\{1,2\}$ denotes the choices of accommodation. 
\\
{\bf \ref{convexr2}}: In \cite{CL01} and also the accommodation example, the reward  $\rewardp$  is chosen as linear in $\action_1$.
This is because $\action_1$ is a deposit made at time 1; so the reward at time 2 is the net wealth minus the deposit at time 1.

Assuming the reward $\rewardp$ to be convex in $\action_1$ is more general and still yields the same  structural result.
\\
{\bf \ref{supermod}} is a 
  supermodularity assumption and implies that
$\action_2$ and $\physical_2$ satisfy Edgeworth  complementarity~\cite{Top98}.
This means that increasing $\physical_2$ increases the marginal value of
choosing $\action_2 = 2$ compared to $\action_2 = 1$. This is intuitive: For the accommodation example, 
 a higher review of $\new$ gives more incentive to choose accommodation $\new$. Supermodularity 
is widely used to characterize the structure of policies in stochastic control and game-theory. 
 By  Topkis' famous theorem \cite{Top98}, supermodularity \ref{supermod} implies  Nash policy $\pol_2^*(\physical_2, \action_1)$ is non-decreasing in $\physical_2$ for fixed $\action_1$. This together with \ref{implicit} implies
that $\pol_2^*$ has a threshold structure  (\ref{eq:pol2threshold}) wrt $\physical_2$ (see proof). In \cite{CL01} and the social media accommodation example,  \ref{supermod} holds trivially  since $\rewardp(\physical_2,\action_2=\known,\action_1)$ is  independent  of $\physical_2$;

To motivate the remaining assumptions, we first note that 
Assumptions \ref{convexr2}-\ref{convexcdf} imply that the anticipatory state $\psych_1$ is convex in $\action_1$ (as shown in the proof).  Since a convex function is maximized at its end points of $\actionspace_1 = [0,1]$, namely 0 and 1, the bang-bang structure (\ref{eq:bangbang})  for $\pol_1^*$ holds.
We now 
 dive deeper into
 \ref{implicit}-\ref{convexcdf}.

 \noindent  {\bf \ref{implicit}}:
 \ref{implicit} is simply an assumption on the well-posedness of the setup; namely, that there exists  a threshold point $ \physical_{2,\thbelief}^*(\action_1) $; implying that the anticipatory agent makes simple intuitive decisions $\action_2$ based on the state $\physical_2$.

\noindent {\bf \ref{convexthreshold}} is a prescriptive assumption on the rewards $ \rewardp$.
From a risk averse point of  view, it is natural that a higher deposit $\action_1$ should result in requiring a substantially higher review $\state_2$ in order for $\action_2$ to forfeit the deposit on $\known$  and switch to $\new$. This is captured by requiring that the threshold  point $\physical_{2,\thbelief}^*(\action_1)$ in \eqref{eq:pol2threshold} is convex in $\action_1$. The natural question then is: What assumptions on the reward guarantee this convexity?
Statement~2 of Theorem \ref{thm:nashstructure} is equivalent to  showing convexity in $\action_1$ of the solution $\physical_2^*(\action_1)$
of 
the algebraic equation  $\diffreward(\physical_2,\action_1) = 0 $. It is here that  \ref{convexthreshold}
is used.
 \ref{convexthreshold} and \ref{implicit}
are sufficient for the
implicit solution to an algebraic equation involving two variables to be convex wrt the other
variable. 
Showing convexity of the implicit solution to an algebraic equation dates back to  \cite{BT66} where \ref{convexthreshold} is used. \ref{implicit} can be relaxed based on the classical implicit function theorem  \cite{Apo74}; see supplementary document for details. In the accommodation example,  \ref{implicit},  \ref{convexthreshold} hold trivially since $\diffreward$ is linear in $\physical_2,\action_1$.

\noindent {\bf \ref{beta}} states that the anticipatory reward is linear in the psychological state.  Therefore $\beta$ denotes the importance of the anticipatory reward relative to the reward at time 2. This assumption is identical to that in \cite{CL01}.

\noindent {\bf \ref{convexcdf}} is also a prescriptive assumption on the system behavior to ensures that the psychological state $\psych_1$ is convex in action $\action_1$.
Actually in \cite{CL01} and the accommodation example, the psychological state $\psych_1$ is linear
and increasing  in action $\action_1$.
From a behavioral point of view, convexity of the psychological state in $\action_1$ is natural since it yields the bang-bang structure (\ref{eq:bangbang}) of the Nash equilibrium which  motivates the ``deliberate avoidance of information behavior'' discussed above.

The convexity of rewards \ref{convexr2} and assumption  \ref{convexcdf} together with Statement 2 imply that  anticipatory (psychological) state $\psych_1$
is convex in $\action_1$. Specifically in the accommodation example and also  \cite{CL01}, 
$\pdfp(\physical_2|\physical_1)$ is uniformly distributed in $\physical_2$,
\begin{equation*}  \begin{split}
  \ant(\action_2=1,\physical_2)&= \physical_2\, I(\state_2 \in [\frac{2+\action_1}
  {3\thbelief},1]) \\  \ant(\action_2=2,\physical_2)&= I(\state_2 \in [0,\frac{2+\action_1} {3\thbelief}])
\end{split}
\end{equation*}
which clearly satisfy \ref{convexcdf}; see the examples for details.

 \subsection{Example 2. Anticipatory Situation Awareness (SA)} \label{sec:sa}
 
 We now discuss an anticipatory decision making example involving Level 3 SA.  The example will be developed further  in the context of quickest time change detection in Sec.\ref{sec:qdnew}.

 \subsubsection{Model}

 The physical states  $\physical_1$ and $\physical_2$ denote the probability that the  threat level of a target (or group of targets)  is $\low\; \threat$ or $\high\;\threat$, at stages 1 and 2. 

 Regarding the actions,
at the first stage the SA system  chooses action
$\action_1 \in [0,1]$ which denotes fraction of resources devoted to tracking a specific target.
At the second stage, the SA makes the final choice of whether to take active measures (e.g.\ intercept the target)  or
choose passive measures (continue to track it),  i.e., $\action_2 \in 
\actionspace_2= \{\activ,\passive\}$. 

Next we model the anticipatory decision making of the SA system.
We choose the anticipatory reward to reflect beliefs about threat levels that will be derived in choosing  respectively,  $\activ$  and $\passive$.
We choose the anticipatory  state $\psych_1$ at time 1 as  the conditional probabilities 
(see (\ref{eq:psych}))
\begin{align}  \psych_1 = \max\{ & 6 \, \pdf(\action_2=\activ, \physical_2 = \high\; \threat| \action_1,\policy_2) , \nonumber \\  & \qquad 4\, \pdf(\action_2 = \passive| \action_1,\policy_2)\} 
  \label{eq:psych_example} \end{align}
(We allocate numerical values  to make the example more readable.)
So the anticipatory reward   increases with the SA's plan to use an active measure if the threat is high.

We now construct the rewards $\reward_1, \reward_2$ defined in (\ref{eq:cl_cost}).
\begin{compactenum}
\item Choosing action $\action_1$ expends  $2 \, \action_1$ resources on planning for $\activ$ measures at time 2.
If $\passive$ is chosen at time 2, then the resources  of $2\,\action_1$ are wasted (lost).
  \item The reward accrued by choosing  $\activ$
when the threat level is $\physical_2$ is $6  \physical_2 \thbelief$; the reward for choosing  $\passive$ is fixed at $4$.
Here\footnote{\label{foot:private}We assume $\private = [\private(1),\private(2)]^\p$ is a 2-dimensional probability vector, i.e., $\beliefdim=2$ in (\ref{eq:thbelief}).  For notational convenience, we refer to $\private(2)$ as $\private$.} $\thbelief\in [0,1]$ is the posterior  probability that the threat level with action $\activ$ is high given information from sensing functionalities.
\end{compactenum}
Based on  the above description, the  rewards are
\begin{equation*}
  \begin{split}
&  \reward_1 = \beta \psych_1, \quad  
 \rewardp(\physical_2,\action_2=\activ, \action_1) = 6\,  \physical_2 \thbelief  - 2\,  \action_1 , \\
& \rewardp(\physical_2,\action_2=\passive,\action_1) = 4
\end{split}
\end{equation*}

\subsubsection{Structure of Nash Equilibrium}   For simplicity, assume $\physical_2$ is uniformly distributed in $[0,1]$. For the above example, we can verify Assumptions   \ref{actionspace}-\ref{convexcdf} hold and therefore Theorem \ref{thm:nashstructure} holds. Specifically, \ref{actionspace} holds by formulation; \ref{convexr2}  holds trivially since $\rewardp$ is linear in $\action_1$; \ref{supermod} holds since $\rewardp(\physical_2,\action_2=\passive,\action_1)$ is  independent  of $\physical_2$; \ref{implicit} and  \ref{convexthreshold} hold trivially since $\diffreward$ is linear in $\physical_2$ and $\action_1$; \ref{beta} holds by construction since it is  easily shown that for optimal policy  $\pol_2^*$,
$z_1 = 4\, \pdf(\action_2 = \passive| \action_1,\policy_2^*)$. Finally,  \ref{convexcdf} holds since $\pdf(\physical_2)$ is the  uniform density by assumption. 

\subsubsection{Consequences}
  Theorem \ref{thm:nashstructure} implies $\pol_2^*$ has a threshold structure (\ref{eq:pol2threshold}), and $\pol_1^*$ has a   bang-bang structure (\ref{eq:bangbang}).
The bang-bang structure  (\ref{eq:bangbang}),   represents a dilemma to the SA system.
The SA system fully utilizes its resources,  $\action_1 = 1$ towards plan $\activ$ if $\beta > \beta^*$.  Yet this  does not guarantee that the SA system  will choose $\action_2 = \activ$ since if $\physical_2 > 
\physical_2^*(\action_1) $, then  the agent will choose $\action_2 =\passive$. Thus a human-in the-loop in the SA system  might deliberately choose not to observe the state $\physical_2$ in order not to lose the effort invested at stage 1.

\subsection{Other Examples}
\label{sec:other-examples}

\textit{Example 3. Airbnb example of Social media accommodation:}  The supplementary document gives  a detailed example in social media accommodation with a similar  dilemma due to the bang-bang Nash equilibrium structure: avoid information at stage 2 so as not to lose the full deposit made at stage 1. 

\textit{Example 4. Asset Prices and Anxiety:} \cite{CL01} presents a two stage model for 
portfolio choice and the anxiety of holding risky assets. The 
anxiety encountered at time 1 depends on the expected reward and variance of the reward at time 2.  

\vspace{0.25cm}
\hrule
\vspace{0.25cm}

{\large {\bf Part 2. Quickest Change Detection for Team  Anticipatory Decision Makers}}

\vspace{0.3cm}

Part 1 of the paper  described how a single anticipatory agent makes decisions over a two-period time horizon.
In Part 2, we
consider a {\em team} of   anticipatory  agents (or equivalently, a single agent that acts multiple times). These anticipatory  agents  interact with each other sequentially and also with a global decision maker to   achieve quickest change detection.
Each anticipatory agent observes the state of nature (Markov chain) in noise  and makes local  decisions as described in Sec.\ref{subsec:clmodel}.
A global decision maker observes these decisions.
\textit{How can a global decision maker use these local decisions  to detect a change in the state of nature}? Specifically the aim
is to 
achieve quickest change detection by minimizing the Kolmogorov-Shiryaev criterion (defined in (\ref{eq:ksd}) below) which  involves the   false alarm and  delay penalties.

\vspace{-0.6cm}

\blue{\subsection*{Examples of Team-based Quickest Detection} Before proceeding  with the quickest change detection formulation, it is helpful to keep the  following examples in mind:\\
(i) {\em Change in Quality of Social Media Accommodation}. Suppose individual anticipatory agents choose  between reserving accommodation in two places. By monitoring these decisions, how can a global decision maker  (e.g. Airbnb)   detect if there is a sudden change in the demand for a specific  accommodation due to the presence of a new competitor (or change on quality in the accommodation)? This example is discussed in the supplementary material as a detailed tutorial.\\
(ii) \textit{Supervisory SA}. Sec.\ref{sec:sa}  discussed the importance of anticipatory situation assessment.  Suppose a supervisory situation assessment (SA) system monitors the  decisions
of individual SA systems. Individual SA systems are anticipatory (as discussed in Sec.\ref{sec:sa})  and monitor an  enemy
target or radar state. How can the supervisory decision maker detect if there is a sudden change in the enemy target (due to a purposeful maneuver)? This example is discussed  in Sec.\ref{sec:teamsa}.\\
(iii)  \textit{Detecting change in betting strategy}.  How to detect a sudden change in the betting strategy of individuals that act sequentially? 
Sec.\ref{sec:spot} discusses a numerical example
which builds on the anticipatory  betting  model of Sec.\ref{sec:bet}.\\
(iv)\textit{ Detecting Market Shocks}. Suppose individual anticipatory  investors make decisions based on their observation of the underlying value of an asset as in Sec.\ref{sec:bet}, where the 
decisions of previous investors affect the individual's belief. How can an analyst detect sudden market shocks? See \cite{AS08} for examples in
 high frequency financial  trading.
}


\section{Anticipatory Quickest Change Detection} \label{sec:qdnew}


{\em Notation.} Since we  consider  the sequential interaction of multiple anticipatory agents,  we adapt  the notation of
Sec.\ref{sec:structure}: 
\begin{compactitem}
\item
 Each anticipatory agent acts  in a predetermined order indexed by $\dtime=1,2,\ldots$.  
\item
   The physical states $\physical_1,\physical_2$ (defined in Sec.\ref{subsec:clmodel}) encountered by agent $\dtime$  are now denoted by $\physical_{\dtime,1},\physical_{\dtime,2}$. 
   \item 
     Anticipatory decisions $\action_1,\action_2$ (characterized in Theorem~\ref{thm:nashstructure}) of agent $\dtime$ are denoted as $\action_\dtime \ole [\action_{\dtime,1}, \action_{\dtime,2}]$.
   \item  The Bayesian belief parameter $\thbelief$  (\ref{eq:rewardparam}) of agent $\dtime$ is $\private_\dtime$. 
   \item Due to the bang-bang structure ((\ref{eq:bangbang}) in Theorem \ref{thm:nashstructure}) of the Nash
     equilibrium policy, $\action_{\dtime,1}$ is independent of $\physical_{\dtime,1}$. Also from (\ref{eq:pol2threshold}), $  \action_{\dtime,2}$   depends on $\physical_{\dtime,2}$ and not $\physical_{\dtime,1}$. So for  convenience we denote $\physical_{\dtime,2}$ as $\physical_\dtime$.  \item The physical state  process $\{\physical_\dtime,\dtime\geq 1\}$ on state space $\physicalstatespace$ is  Markovian  with transition density $\pdf(\physical_{\dtime+1}| \physical_\dtime,\state_\dtime)$, see  (\ref{eq:rewardparam}).
     Here $\state_\dtime$ is the state of nature process (defined below)
     that models the jump change  we aim to detect.  
\end{compactitem}

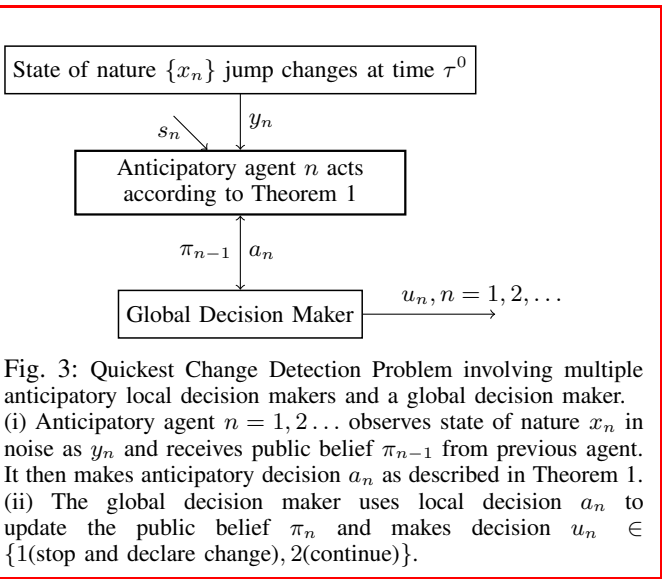
\begin{figure}[h] 
  \centering
  \fbox{ \begin{minipage}{8.5 cm} \mbox{} \vspace{0.4cm} \\
      \begin{tikzpicture} [node distance =1.5cm and 2cm, auto]
                  \tikzset{every node}=[footnotesize=\small]
                     \node[blockb](nature){State of nature $\{\state_\dtime\}$ jump changes at time $\tau^0$};
                  \node [blockf,below of=nature] (l2) {Anticipatory agent $\dtime$ acts according to   Theorem~\ref{thm:nashstructure}};
                 
    \node [blockb,below of=l2,node distance=1.75cm] (global) {Global Decision Maker};
    \node [above of=l2,node distance=1.3cm] (y2)[draw=none]{};
    \node [right of=global,node distance=3.5cm] (globalaction)[draw=none]{};
    \node [above of=l2,left of =l2,node distance=1cm] (s2)[draw=none]{};
    %
      \draw[->](s2) --  node[left]{$\physical_\dtime$}   (l2);
    \draw[->](y2) -- node[above,right]{$\obs_\dtime$}  (l2);
       \draw[->](global) -- node[above,pos=0.9]{$\globalaction_\dtime,\dtime=1,2,\ldots$}  (globalaction);
    \draw [<->] (l2) -- node[above,right]{$\action_\dtime$} node[below,left]{$\belief_{\dtime-1}$} (global);
   \end{tikzpicture}  

   \caption{\small Quickest Change Detection Problem involving multiple anticipatory local decision makers and a global decision maker. \newline
     (i)  Anticipatory agent   $\dtime=1,2\ldots$ observes  state of nature $\state_\dtime$ in noise as $\obs_\dtime$ and  receives public belief $\belief_{\dtime-1}$ from previous agent.
     It then makes anticipatory decision $\action_\dtime$ as described in
     Theorem~\ref{thm:nashstructure}. \newline (ii)  The global decision maker uses local decision $\action_\dtime$ to  update the  public belief $\belief_\dtime$ and makes decision $\globalaction_\dtime \in \{1 \text{(stop and declare change)}, 2 \text{(continue)}\}$. 
  }
  \label{fig:schematic2}
\end{minipage}
}
\end{figure}

{\em Jump Change Model}. \blue{The state of nature   $\{\state_n\in \{1,2\}, n \geq 0\}$  models the change event we aim to detect.}  It starts state 2  at time $0$
and jumps to  state 1  at  a geometrically distributed  random time $\changetime$ with mean $1/(1-\param)$, for some prespecified $\param \in [0,1)$.
Equivalently,  $\{\state_\dtime\}$ is  a 2-state Markov chain  with
absorbing  transition matrix and initial  probability
\beq \tp  =  \begin{bmatrix} 1 & 0 \\ 1-\param & \param 
\end{bmatrix} , \quad \belief_0 = \begin{bmatrix} 0 \\ 1 
\end{bmatrix}
\label{eq:tp} \eeq
with  change time
$ \changetime = \inf\{\dtime: \state_\dtime = 1\}$.  
Clearly the transition matrix $\tp$ implies that
$\E\{\tau^0\} = 1/(1-\param)$.


\vspace{-0.15cm}

\subsection{Multi-agent Quickest Detection Protocol} \label{sec:protocol}
Quickest  detection involves detecting  change time $\tau^0$ with minimal cost.  The multi-agent formulation considered here 
comprises of interacting local decision makers (anticipatory agents) and a   global decision maker (see Figure \ref{fig:schematic2}):
\begin{compactenum}
\item The jump change process (state-of-nature) $\{\state_\dtime,\dtime\geq 0\}$ affects the transition kernel and reward of  the  physical state  process  $\{\physical_\dtime,\dtime\geq 1\}$; see~(\ref{eq:rewardparam}).
\item Each anticipatory agent  $\dtime$ acts sequentially  indexed by $\dtime=1,2,\ldots$. Agent $\dtime$ observes state of nature $\state_\dtime$ in noise and makes a local decisions $\action_\dtime=(\action_{\dtime,1},\action_{\dtime,2})$ corresponding to actions $\action_1, \action_2$ in Sec.\ref{sec:structure}.

\item 
Based on the history of local actions $\action_1,\ldots,\action_\dtime$, the global decision maker chooses action  $$\globalaction_\dtime \in \{ 1 \text{ (stop and announce change)}, 2 \text{ (continue)} \}$$
\end{compactenum}

Define the public belief $\belief_\dtime$ and private belief $\private_\dtime$   at time $\dtime$ as the posterior
distributions initialized with 
$    \private_0 = \public_0 = \begin{bmatrix} 0 & 1  \end{bmatrix}^\p$:
\begin{equation} \begin{split}  \public_\dtime(\state) &= \prob(\state_\dtime = \state | \action_1,\ldots,\action_\dtime), \quad \state = 1,2.  \\
    \private_\dtime(\state) &= \prob(\state_\dtime = \state | \action_1,\ldots,\action_{\dtime-1},\obs_\dtime),
   \label{eq:public} 
\end{split}
\end{equation}
where $\obs_\dtime$ is the private observation recorded by agent $\dtime$
(see (\ref{eq:oprob}).
Note $\private=[1- \private(2), \private(2)]^\p$ and  $\public= [1-\public(2), \public(2)]^\p$; they lie in the one dimensional simplex $\Belief = [0,1]$.

We are now ready to describe the 
 multi-agent quickest detection protocol, see also Figure \ref{fig:schematic2} for a   schematic setup.   

\vspace{0.5cm}\hrule height.8pt depth0pt \kern2pt
{\bf Protocol 1. Multi-Agent Bayesian  Quickest Detection}
\hrule height.8pt depth0pt \kern2pt
\begin{compactenum} \item \label{item:local} \textit{Local anticipatory decision maker $\dtime$} 
\begin{compactenum} 
\item \label{item:obtain} Obtains  public belief $\public_{\dtime-1}$  from global decision maker.
\item \label{item:obs} The agent  records  private noisy observation $\obs_\dtime\in \obspace$ of state of nature $\state_\dtime$  with
  conditional density
  \beq \oprob_{\state,\obs} = \pdf(\obs_\dtime=\obs|\state_\dtime=\state) \label{eq:oprob} \eeq
\item \label{item:private} {\em Private Belief}. The agent evaluates the  private belief 
\begin{align}
  \private_\dtime &= \filter(\public_{\dtime-1},\obs_\dtime) \label{eq:privateupdate}
  \text{ where, }  \filter(\public,\obs) = \frac{\oprob_\obs \, \tp^\p \public}{\filterd(\public,\obs)}, \\ & 
  \filterd(\public,\obs) = \ones^\p \oprob_{\obs} \tp^\p \public , \quad
  \oprob_\obs = \diag(\oprob_{1,\obs},\oprob_{2,\obs}) \nn
\end{align}
\item {\em Change Event \&  Local decision}. \label{item:ld} \blue{The agent's private belief $\private_\dtime$   affects its reward and transition kernel of   physical state  process  $\{\physical_\dtime,\dtime\geq 1\}$ as in~(\ref{eq:rewardparam}):
\beq
\begin{split}
\rewardp(\physical_2,\action_2,\action_1) &= \sum_{\state \in \statespace} \reward_2(\physical_2,\action_2,\action_1,\state)\, \private(\state) \\
\pdfp(\physical_2|\physical_1) &= \sum_{\state \in \statespace} \pdf(\physical_2|\physical_1,\state)\, \private(\state)
\end{split}
\tag{\ref{eq:rewardparam} repeated}
\eeq
  } \\ The agent uses $\thbelief_\dtime$, $\physical_\dtime$  to make anticipatory decisions $\action_\dtime=(\action_{\dtime,1},\action_{\dtime,2})$ via
  (\ref{eq:bangbang}), (\ref{eq:pol2threshold})  in Theorem~\ref{thm:nashstructure}.
\end{compactenum}
\item {\em Global decision maker}.   \label{item:global} Based on the decisions  $\action_{\dtime}$ of local decision maker  $\dtime$, the global decision maker:
  \begin{compactenum} \item \label{item:public} Updates the public belief  from $\public_{\dtime-1}$ to $\public_{\dtime}$ as
    \begin{align}
 & \public_\dtime = \filterg(\public_{\dtime-1},\action_\dtime,\physical_\dtime) \label{eq:pubupdate} \\
  &  \filterg(\public,\action,\physical) = \frac{\oprobg_\action(\physical) \, \tp^\p \public}{\filterdg(\public,\action,\physical)}, \quad
    \filterdg(\public,\action,\physical) = \ones^\p \oprobg_{\action}(\physical) \tp^\p \public \nn   \\
            & \text{ where }   \oprobg_\action(\physical) = \diag(\oprobg_{1,\action}(\physical),\oprobg_{2,\action}(\physical)), \nn \\
&\oprobg_{\state,\action_n}(\physical) = \prob(\action_\dtime=\action|\state_n=\state,\public_{\dtime-1},\physical_\dtime = \physical)
                   \label{eq:actionprob}
\end{align}
The action probabilities  $\oprobg_{\state,\action}$ are computed as
\beq \oprobg_{\state,\action}(\physical) = \int_\obspace I(\polpT(\physical,\action_{\dtime,1}) = \action_{\dtime,2}) \oprob_{x,y} dy \label{eq:oprobgcompute}
\eeq
Recall $a_n=(a_{n,1},a_{n,2})$ and $\polp2$ is the local decision maker's subgame Nash equilibrium policy~(\ref{eq:pol2threshold}).

\item \label{item:globalaction} Chooses global action $\globalaction_\dtime$ using optimal  policy $\globalpolicy^*$:
  \begin{multline}  \label{eq:globalDM}
  \globalaction_\dtime = \globalpolicy^*(\public_\dtime,\physical_\dtime)  \in \{1 \text{ (stop)}, 2 \text{ (continue)}\}.  
\end{multline}
\item If $\globalaction_\dtime =2 $, then 
    set $\dtime$ to $\dtime+1$ and go to Step~1. \\ If $\globalaction_\dtime =1$, then  stop and announce change.
  \end{compactenum}

\end{compactenum}
  \kern4pt  \hrule width 0.4\textwidth \kern4pt 


  \blue{{\em Remark}. The reader should note that there are two states in our formulation,
    namely, the state of nature $\state$ (that jump changes)  which  is observed in  noise by anticipatory agents,
and the physical state $\physical$ which  determines the agent's  anticipation. As specified  in Step 1d,  the state of nature $\state$  affects  the transition kernel of $\physical$ and reward of each  anticipatory agent.} 

\vspace{-0.4cm}
  
\subsection{Quickest Detection Objective of Global Decision Maker} \label{sec:gdm}
  We assume the global decision maker knows $\tp$ (\ref{eq:tp}), physical state $\physical_\dtime$, agent's action $\action_\dtime$,
  and agent's policy $\polp2$.
  The global decision maker does not know $\obs_\dtime$ (agent's observation/perception) or the agent's private belief~$\private_\dtime$
  in Step 1.
  For simplicity, we  assume  all agents have the same anticipatory model parameters; otherwise the optimal quickest detection strategy is non-stationary. We emphasize that the transition probabilities of the physical state and utility  of each agent depends on its private belief $\private_\dtime$ of $\state_\dtime$, see (\ref{eq:rewardparam}) in Protocol 1.
  
  The aim of quickest detection is to determine the jump time $\tau^0$ of the state of nature $\{\state_\dtime\}$, i.e., evaluate the optimal stationary policy $\globalpolicy^*$ of the global decision maker  that  minimizes   the Kolmogorov--Shiryaev
  criterion for detection of disorder~\cite{Shi63}:
  \begin{equation} \begin{split}
 \globaltogo_{\globalpolicy^*}(\belief,\physical) &= \inf_{\globalpolicy} \globaltogo_\globalpolicy(\belief,\physical), \\
 \globaltogo_\globalpolicy(\belief,\physical) &=   d \, \E_{\globalpolicy}\{(\tau - \tau^0)^+\} +  f \,\P_\globalpolicy(\tau < \tau^0) .
\end{split}
\label{eq:ksd} 
\end{equation}
Here  $\tau = \inf\{\dtime: \globalaction_\dtime= 1\}$ is the time at which the global decision maker announces the change.
The  parameters $d$ and $f$ specify the delay penalty and false alarm penalty, respectively. So waiting too long to announce a change incurs a delay penalty $d$ at each time instant after the system has changed, while declaring
a change before it happens, incurs a false alarm penalty $f$.  $\prob_\globalpolicy$ and $\E_\globalpolicy$ are the probability measure and expectation
of the evolution of the local decisions, observations  and Markov state which are strategy dependent. In (\ref{eq:ksd}), $\belief$ denotes the initial distribution of the Markov chain $x$ and $\physical$ is the initial state of the physical state process.

  \textit{Remark. Comparison with Classical Quickest Detection}.
 Quickest detection with anticipatory agents  (Protocol~1)  is substantially more general
  than classical quickest detection.

As shown in \eqref{eq:table}, in classical quickest detection the decision maker has access to observations $\{\obs_\dtime\}$ which are noisy measurements of $\{\state_\dtime\}$, and then computes belief $\private_\dtime$. In comparison, in our framework
 the global decision maker only has access to the local decisions $\{\action_\dtime\}$ of the anticipatory agents; these local decisions  depend on $\obs_\dtime$ via the dynamics in Steps 1c and 1d in Protocol 1. 
In particular, the public belief $\public_\dtime$ in
(\ref{eq:actionprob}) depends on the action likelihoods; whereas in classical quickest detection the belief depends on the observation likelihoods.
The objective of classical quickest detection
is exactly the Kolmogorov--Shiryaev
  criterion for detection of disorder (\ref{eq:ksd}) except that the belief is the classical Bayesian posterior $\pdf(\state_\dtime|\obs_{1:\dtime})$ instead of $\public_\dtime$ defined in (\ref{eq:public}).

 \begin{figure} 
  \begin{minipage}[t]{3.9cm}
    \begin{equation*}   \begin{split}
       & \text{\bf Anticipatory} \\ \hline
       \state_n &\sim \tp \text{ (change state)} \\
        & \hspace{-0.5cm} \text{Local Anticipatory  Decision:} \\
      \obs_n &\sim \oprob_{\state_n, \obs} \text{ (observation)} \\
      \private_{\dtime} &= \filter(\public_{\dtime-1}, \obs_{\dtime})\\
      \action_{n} &=   \mu^*_{2,\private_{n}}(\physical_{n,2},\action_{\dtime,1}) \\
     & \hspace{-0.5cm} \text{Global Decision maker:} \\
      \belief_{n} &= \filterg(\belief_{n-1},\action_{n},\physical_{\dtime}) \\
      \globalaction_{n} &= \globalpolicy^*(\belief_{n})\in \{1,2\}
    \end{split}
  \end{equation*}
\end{minipage}
\hfill \vline \hfill
\begin{minipage}[t]{4.5cm}
  \begin{equation}   \begin{split}
       & \text{\bf Classical} \\ \hline
       \state_n &\sim \tp \text{ (change state)}\\
        & \hspace{-0.5cm} \text{Decision maker:} \\
      \obs_n &\sim \oprob_{\state_n, \obs} \\
      \private_{\dtime} &= \filter(\private_{\dtime-1}, \obs_{\dtime})\\
      \globalaction_{n} &= \globalpolicy^*(\private_{n}) \in \{1,2\}
    \end{split} \label{eq:table}
  \end{equation}
\end{minipage} 
\end{figure}

\vspace{-0.5cm}

\subsection{Example.  Change Detection in Team Situation Awareness} \label{sec:teamsa}
  Sec.\ref{sec:sa} described  an individual anticipatory situation awareness (SA) system.
 In complex environments 
 individual SA is no longer  adequate.
We consider here \textit{ team-level} SA \cite{GCW06}.  For example, \cite{EHF10}  introduced a situational
adapting system to assess team SA for fighter pilots based
on information fusion. To achieve team SA, individual pilots
need to develop and retain their own SA while performing the
task, share their SA and notice relevant activities of other
members in the team. In the simplest sequential framework of Team SA, we have the setup of Protocol 1 where:
\begin{compactenum}
\item The underlying state of nature $\state_\dtime$ denotes  the enemy  target or radar state  that is  monitored 
  by the  SA system.
    \item $\obs_\dtime$ are  measurements  of  the enemy's state   $\state_\dtime$. 
    \item
      $\public_{\dtime-1}$ is the  enemy's  belief  $\pdf(\state_{\dtime-1}|\action_1,\ldots,\action_{\dtime-1})$ obtained from a Bayesian tracking algorithm.
    \item  The physical state $\physical_\dtime$ is  the  probability of  threat. Its transition kernel 
    is modulated by ground truth $\state_\dtime$ (\ref{eq:rewardparam}).
\item Individual agents in the team  SA agent  make decisions  $\action_{\dtime,1}, \action_{\dtime,2}$  according to Protocol 1   and  relay them to subsequent SA systems in the team.
\end{compactenum}
Then quickest  detection is motivated as  follows:
by monitoring the   decisions $\{\action_\dtime\}$ of the individual SA systems, how can a supervisory system  detect if there is a 
sudden change in the state $\{\state_\dtime\}$? 
Such a  change  is reflective of the enemy target making purposeful  maneuvers; or the enemy radar  switching modes between  search, acquisition or  track. Since it operates at a higher level of abstraction,  the supervisory system  does not  have access to the observations $\obs_\dtime$ of individual SA systems.

 A similar framework in social media accommodation is discussed in the supplementary document.
Given the sequence of decisions $\{\action_\dtime\}$, the
 global decision maker (e.g. Airbnb)  wishes to detect if there is a 
 sudden appearance of competition or sudden change in quality of the accommodation $\state_\dtime$. The physical state $\physical_\dtime$ is the probability of a good review (review histogram) and its kernel depends on the ground truth  $\state_\dtime$.

\subsection{Discussion of Protocol 1}

{\em 1. Sensor-human Interface}. 
Suppose each anticipatory human decision maker is equipped with a sensing/computing device
that  performs   Steps \ref{item:obtain} to \ref{item:private}. Specifically, the noisy observation $\obs_\dtime$ in
Step \ref{item:obs}
  is obtained by a sensor/computing device  which  then uses Bayes rule  to evaluate the private belief $\private_\dtime$ in Step \ref{item:private} according to (\ref{eq:privateupdate}).
The sensing functionality then  provides $\private_\dtime$ to   the  anticipatory  decision maker.
Recall that 
  $\private_\dtime$  enters the parametrized rewards of the anticipatory decision maker as discussed in (\ref{eq:thbelief}). Finally, the anticipatory decision  maker chooses action $\action_\dtime$ in Step \ref{item:ld} according to the framework  in Sec.\ref{sec:structure}.
Thus Step \ref{item:local} preserves the  simplicity of the anticipatory  human decision making model in  \cite{CL01}.

  {\em 2. Global decision maker}. Step \ref{item:global} details the decision making framework of the global decision maker. The global decision maker has access to the physical state $\physical_\dtime$ and the actions $\action_{\dtime,1}, \action_{\dtime,2}$  of the local decision maker. These are used by the global decision maker in Step \ref{item:public} to update the public belief in (\ref{eq:pubupdate}). The action likelihoods in  (\ref{eq:oprobgcompute})
  follow from (\ref{eq:privateupdate}) and the fact that
  $$  \oprobg_{\state,\action} (\physical)= \int I(\polp2(\physical,\action_{\dtime,1}) = \action_{\dtime,2}) \,\pdf(\private|\public_{\dtime -1} ,\obs) \oprob_{x,y}\, d\private\, d\obs$$
Finally in Step \ref{item:globalaction},  the global decision maker applies the optimal policy $\globalpolicy^*$ to the updated public belief $\public_\dtime$, to  choose
  whether to continue or stop (announce change).

\blue{{\em 3. Information Structure}.
  Protocol 1 depicts three types of  interactions.  Local decision makers  learn from previous local decision makers. Second, the local decisions $\action_\dtime$ determine  global decisions $\globalaction_\dtime$.
  Finally, if the global decision maker chooses $\globalaction_\dtime = 2$, then  the protocol  continues to the next time; otherwise a change is detected and the process stops.}
  
  {\em 4. Comparison with Bayesian social learning}.  Protocol 1 generalizes classical Bayesian social learning \cite{Cha04} in two ways. First,
  the public belief update (\ref{eq:pubupdate})  is a generalization of the Bayesian social learning filter \cite{Kri16}, where the local decision maker is a myopic optimizer (in comparison,  we now have a two-stage anticipatory local decision maker). Second, the local decision makers operate in closed loop; they are controlled by the global decision maker.


\subsection{Stochastic Dynamic Programming Formulation} \label{sec:dp}
The aim of this section is to formulate the global decision maker's quickest  change detection policy $\globalpolicy^*(\belief,\physical)$  (defined in (\ref{eq:ksd})) as the solution of a stochastic dynamic programming equation.
The quickest detection problem (\ref{eq:ksd}) is an example of a stopping-time partially observed Markov decision process (POMDP)  problem with a stationary optimal policy \cite{Kri16}.

\subsubsection{Costs} To present the dynamic programming equation, as is standard,
we first  formulate the false alarm and delay costs (\ref{eq:ksd})  incurred  by the global decision maker in terms of the public belief  (also called the information state), see \cite{Kri16}.

(i) {\em  False alarm penalty}:  
If global decision $\globalaction_\dtime=1$ (stop)  is chosen at time $\dtime$, then the Protocol 1 terminates.  If  $u_\dtime=1$   is chosen before  the change point $\tau^0$, then a false
alarm penalty is incurred.
The false alarm event  $ \{\state_\dtime=2, \globalaction_\dtime = 1\}$ represents the event
that a change is announced before the change happens at time $\tau^0$.
Recall (\ref{eq:tp})  the jump change occurs at time $\tau^0$ from state 2 to state 1.
\blue{Then recalling $f\geq 0$ is the false alarm penalty in \eqref{eq:ksd}, the expected   false alarm penalty~is
$$ f \,\P_\globalpolicy(\tau < \tau^0)  = f\,\E_\globalpolicy \{\E\ I(\state_\dtime=2,\globalaction_\dtime=1)| \G_\dtime\}\}$$ 
\beq
  \G_n = \sigma\text{-algebra generated by } (\action_1,\ldots,\action_\dtime)  \label{eq:sigalg}\eeq
 Clearly
  $\E\ I(\state_\dtime=2,\globalaction_\dtime=1)| \G_\dtime\}$
  can be expressed in terms of  public belief $\belief_\dtime(2) = P(\state_n=2| \action_1,\ldots,\action_\dtime)$ as}
\beq \Cost(\public_{\dtime},\globalaction_\dtime=1) =  f \,e_2^\p \public_\dtime , \quad
\text{ where }
e_2 = [0\quad 1]^\p
. \label{eq:cp1}\eeq

(ii) {\em Delay cost of continuing}: If global decision $\globalaction_\dtime=2$ is taken then Protocol 1   continues to the next time.
A delay cost is incurred when the event  $ \{\state_{\dtime} = 1, \globalaction_\dtime = 2\}$ occurs,
i.e., no change is declared at time $\dtime$, even though the state has changed at time $\dtime$.
The expected delay cost is $d\,\E\{I(\state_{\dtime} = 1, \globalaction_\dtime=2) | \G_\dtime\}$ where
$d > 0$ denotes the delay cost. In terms of the public belief, the delay cost is
 \beq  \Cost(\public_{\dtime},\globalaction_\dtime=2) = d e_1^\p \public_\dtime, \quad
\text{ where }
e_1 = [1 \quad 0]^\p . \label{eq:cp2} \eeq

We can   re-express  Kolmogorov-Shiryaev criterion (\ref{eq:ksd})
as\footnote{The formal construction is as follows. Let $(\Omega,\mathcal{F})$ denote the underlying measurable space where
  $\Omega = (\statespace \times \globalactionspace \times \obspace \times \physicalstatespace)^\infty$ is the product space endowed the with product topology, and $\mathcal{F}$ is the corresponding $\sigma$-algebra. Then for any $\belief \in \Belief$, $\physical \in \physicalstatespace$ and policy
stationary policy $\globalpolicy$, there exists a unique probability measure $\prob_{\globalpolicy}$ on  $(\Omega,\mathcal{F})$, see \cite{HL96}. In (\ref{eq:ksd}) and (\ref{eq:beliefcost}), $\E_\globalpolicy$ denotes the expectation wrt measure $\prob_\globalpolicy$.}
\beq
\globaltogo_\globalpolicy(\belief,\physical) =
\E_\globalpolicy\{ \sum_{\dtime=0}^{\tau-1} \Cost(\belief_\dtime,2) + \Cost(\belief_\tau, 1) \}
\label{eq:beliefcost}
\eeq
where $\tau = \inf\{\dtime: \globalaction_\dtime = 1\}$ is adapted to the $\sigma$-algebra $\G_\dtime$.
Since $\Cost(\belief,1)$, $\Cost(\belief,2)$ are non-negative and bounded for $\belief \in \Belief$, stopping is guaranteed in finite time. 

\subsubsection{Bellman's equation for  Quickest Detection Policy}
Consider  the costs (\ref{eq:cp1}), (\ref{eq:cp2}) defined in terms of the public belief $\public$. Then
the optimal stationary policy
$\globalpolicy^*(\public,\physical)$  defined  in (\ref{eq:globalDM}),  (\ref{eq:ksd}).
and associated value function
 $\valuef(\public,\physical) $ 
are the solution of 
 Bellman's dynamic programming  functional equation \cite{Kri16}
 \begin{equation} \label{eq:bellman}
   \begin{split}
   \Qfn(\public,\physical,1) &\ole \Cost(\public,1), \\
\Qfn(\public,\physical,2) &\ole \Cost(\public,2)
\\ & \hspace{-1.1cm}  \;\; +\int_\physicalstatespace  \sum_{\action \in \actionspace_1\times \actionspace_2} \!\!\!\!\! \valueg\left( \filterg(\public ,\action,\bar{\physical}) , \bar{\physical}\right) \filterdg(\public,\action,\bar{\physical}) \, \pdf(\bar{\physical}|\physical)\} \, d\bar{\physical}
   \\
\globalpolicy^*(\public,\physical)&= \arg\min\{\Qfn(\public,\physical,1),\Qfn(\public,\physical,2)\}, \\
 \valueg(\public,\physical) &= \min \{ \Qfn(\public,\physical,1),\Qfn(\public,\physical,2)\} = \globaltogo_\globalpolicy^*(\public,\physical)  
\end{split}
\end{equation}
The public belief update $\filterg$ and normalization measure $\filterdg$ were defined in
(\ref{eq:pubupdate}).  Recall  (\ref{eq:globalDM}) that $\globalaction_\dtime = \globalpolicy^*(\belief_\dtime,\physical_\dtime)$ is the global decision maker's action whether to continue or stop.

The goal of the global decision-maker is to solve for the optimal quickest change  policy $\globalpolicy^*$ in (\ref{eq:bellman}) or equivalently, determine the  optimal stopping set  $\Stop$
\beq
\Stop =  \{\public,\physical : \globalpolicy^*(\public,\physical) = 1\} = \{\public,\physical:  \Qfn(\public,\physical,1) \leq \Qfn(\public,\physical,2)\}
 \label{eq:stopset} \eeq


\subsubsection{Value Iteration Algorithm}
 
 The optimal policy  $\globalpolicy^*(\public,\physical)$ and value function $\valueg(\public,\physical)$ can be constructed as  the solution of a fixed point iteration of Bellman's equation (\ref{eq:bellman}) -- the resulting algorithm is called the value iteration algorithm.
The value iteration algorithm  proceeds as follows: Initialize $\valueg_0(\public,\physical) = 0 $ and for iterations $k=1,2,\ldots$
 \begin{equation} \begin{split}
\valueg_{k+1}(\public,\physical) &= \min_{u \in \globalactionspace} \Qfn_{k+1}(\public,\physical,\globalaction), \\
\globalpolicy^*_{k+1}(\public,\physical)&= \argmin_{u \in \globalactionspace} \Qfn_{k+1}(\public,\physical,\globalaction) \quad
               \public \in \Belief,\\
\Qfn_{k+1}(\public,\physical,1) &= C(\public,1) , \quad  \Qfn_{k+1}(\public,\physical,2) =  C(\public,2) \\
& \hspace{-1.7cm} +\int_\physicalstatespace \sum_{\action \in \actionspace_1\times\actionspace_2}  \valueg_k\left( \filterg(\public,\bar{\physical},\action) ,\bar{\physical}\right) \filterdg(\public,\bar{\physical},\action)\, \pdf(\bar{\physical}|\physical) d\bar{\physical},
\end{split}
\label{eq:vi}
\end{equation}
Let $\mathcal{B}$ denote the set of bounded real-valued functions on $\I$.
For any $\valueg,\tilde{\valueg} \in \mathcal{B}$ and $\public \in \I$,  define the sup-norm  metric
$\sup\|\valueg(\public,\physical) - \tilde{\valueg}(\public,\physical)\|$,  $\physical \in \physicalstatespace$. 
Since $\Cost(\public,1)$, $C(\public,2)$, $\public \in \I$, are  bounded, 
the value iteration algorithm (\ref{eq:vi})   generates a  sequence of lower semi-continuous value functions
$\{\valueg_k\} \subset \mathcal{B}$ that converge pointwise
as $k\rightarrow \infty$ to $\valueg(\public,\physical) \in \mathcal{B}$, the solution of Bellman's equation \cite{HL96}.

\noindent {\bf Summary}. Protocol 1 describes the quickest detection protocol involving  anticipative agents acting sequentially. Each local decision maker (agent) $\dtime=1,2,\ldots$ makes  anticipatory decisions $\action_{\dtime,1},\action_{\dtime,2}$ according to the framework in Sec.\ref{sec:structure}. The global decision maker uses these actions to make decision $\globalaction_\dtime = \globalpolicy^*(\belief_\dtime,\physical_\dtime)  \in \{1,2\}$.  The 
optimal detection policy   $\globalpolicy^*$ of the global decision maker satisfies Bellman's equation (\ref{eq:bellman}) and can be constructed by  value iteration algorithm (\ref{eq:vi}).

\blue{Classical quickest detection is a special case of
   (\ref{eq:bellman}), (\ref{eq:stopset}) with $\Qfn(\public,\physical,\globalaction)$
   independent of $\physical$, $ \pdf(\bar{\physical}|\physical)\} = I(\bar{\physical} = \physical)$, and  belief $\belief$ replaced by  classical Bayesian update~(\ref{eq:table}). In classical quickest detection the optimal policy has a threshold structure and the stopping region $\stopset$ is convex; however, these properties do not hold for the multi-agent case considered here.
}

\section{Structural Results for Quickest Detection with Anticipatory Agents} \label{sec:structure2}

The previous section formulated Bellman's  dynamic programming equation for the quickest detection  policy of the global decision maker. However,
since  the belief  space $\Belief$ in (\ref{eq:thbelief}) is a unit simplex (space of probability vectors),  the value iteration algorithm  (\ref{eq:vi})
does not directly
yield a practical solution for computing stopping set $\Stop$  since 
$\valueg_k(\belief) $ needs to be evaluated on the continuum $\belief \in \Belief$. Specifically, in quickest detection, since $\state_k \in \{1,2\}$, the belief space 
$\Belief$ is a 1-dimensional simplex comprising 2-dimensional beliefs of the form $\belief = [1-\belief(2), \belief(2)]^\p$. The value iteration algorithm (\ref{eq:vi})  can be solved numerically  by one-dimensional grid discretization of~$\Belief$.

The aim of this section is to characterize mathematically the structure of the belief updates and achievable optimal cost in quickest detection without brute force computations. 

Specifically we discuss 5 important structural results below:
\begin{compactenum} \item The private belief update of individual anticipatory  agents  follows simple rules justifying  human decision-making.
\item Even though the public belief update depends on the action probabilities $\oprobg$ (\ref{eq:actionprob}) where $\belief \in \Belief$ is continuum, there are only a finite number of such  action probabilities.
\item In stark contrast to classical quickest detection, the value function (\ref{eq:bellman}) in Bellman's equation for quickest detection with anticipative agents is not necessarily concave.
 \item We give numerical examples of the optimal quickest detection policy to highlight the unusual structure of non-concave value function  and non-convex stopping regions. Our numerical examples illustrate change-blindness and detecting a change in betting strategy.
\item Finally, by using Blackwell dominance, we show that the cumulative cost incurred is always larger than classical quickest change detection.
 
\end{compactenum}

\subsection{Private Belief Update follows simple monotone rules} As discussed at the beginning of Sec.\ref{sec:qdnew}, the agent either uses a sensing/computing device to evaluate its private Bayesian belief  or constructs an approximation  to the private belief in order to make an anticipative  decision.
 Below we show that the Bayesian update for the private belief is monotone in the observation and prior; thus it follows simple rules and is  a useful idealization of human decision making.

Recall  Theorem \ref{thm:nashstructure} asserted monotonicity of the anticipatory decision maker's policy
$   \polp2(\physical_2,\action_1)$ wrt physical state $\physical_2$. Here we show monotonicity wrt the Bayesian parameter $\belief $ (recall $\belief $ is the prior for $\private$ in the Bayesian update (\ref{eq:privateupdate})) and observation $\obs$.
We make the following assumptions
 \begin{enumerate}[label=(A{\arabic*}),resume]
\item  \label{itemtp2} The observation likelihoods $\oprob_{\state,\obs}$ (\ref{eq:oprob}) are TP2 (totally positive of order 2); that is,
  $\oprob_{\Bar{\state},\obs}B_{\state,\bar{\obs}} \leq B_{\state,\obs} B_{\bar{\state},\bar{\obs}}$,
  $\bar{\state} > \state$, $\bar{\obs} > \obs$.
\item \label{supermod2} $\reward_2(\physical_2,\action_2,\action_1,\state)$ (see (\ref{eq:rewardparam})) is supermodular in $(\state,\action_2)$, i.e., 
  $\reward_2(\physical_2,\action_2,\action_1,\bar{\state}) - \reward_2(\physical_2,\action_2,\action_1,{\state}) $ is increasing in $\action_2$.
\end{enumerate}

\ref{itemtp2}  is widely studied in monotone decision making; see the classical paper  \cite{KR80}; numerous examples of noise distributions are  TP2. As described in
\cite{Mil81}, observation $\bar{\obs}$ is said to be more ``favorable news''  than observation $\obs$ if \ref{itemtp2}  holds. \ref{supermod2}
 is a supermodularity condition on the rewards; see \ref{supermod}.

In the theorem below recall that $ \optpolicy_{2,\filter(\belief,\obs)}$ is the subgame Nash equilibrium of the local anticipatory decision maker. 
\begin{theorem} \label{thm:monotone}
 The following properties hold for the  anticipatory action $\action_{\dtime,2} = \optpolicy_{2,\filter(\belief,\obs)}(\physical,\action_{\dtime,1})  $ in  (\ref{eq:pol2threshold}) made by agent $n$:
\begin{compactenum}
\item
Under  \ref{itemtp2} and \ref{supermod2},   $\action_{\dtime,2} $  is increasing and ordinal  in observation~$\obs$. That is for any monotone function $\phi$, it follows that  $ \phi(\action_{\dtime,2})$ is also increasing in $\obs$.
\item Under \ref{itemtp2},  $\optpolicy_{2,\filter(\belief,\obs)}(\physical,\action_{\dtime,1})  $  is increasing in belief $\belief$ with respect to the monotone likelihood ratio (MLR) stochastic order\footnote{ \label{footnotemlr} Given probability mass functions
$\{p_i\}$ and $\{q_i\}$, $i=1,\ldots,X$ then
$p$ MLR dominates $q$ if  $\log p_i - \log p_{i+1} \leq \log q_i - \log q_{i+1}$.}  for any observation~$y_n$.\qed
\end{compactenum} 
\end{theorem}

We can interpret Theorem \ref{thm:monotone} as follows. If anticipative agent $ \dtime $  makes recommendations that are monotone and ordinal in the observations and monotone in the prior, then they mimic
the Bayesian social learning model.  Even if the agent does not exactly follow a Bayesian social learning model,  its monotone ordinal behavior implies that such a Bayesian model is  a useful idealization. 
Humans typically make {\em monotone} decisions - the more favorable the private  observation,
the higher the recommendation. Humans   make {\em ordinal} decisions\footnote{Humans typically convert numerical attributes to ordinal scales before making  decisions. For example,
it does not matter if the cost of a meal at a restaurant is \$200 or \$205; an individual would classify this cost as ``high". 
Also credit rating agencies use ordinal symbols such as AAA, AA, A.} since humans tend to think in symbolic ordinal terms.

 We now  discuss assumption \ref{supermod2}.
 Denote the reward vector $$\reward_{a}\ole [\reward_2(\physical_2,\action_2=a,\action_1,\state=1),\ldots, \reward_2(\physical_2,\action_2=a,\action_1,\state=\beliefdim)]^\p$$
 Then \ref{supermod2}  is a stronger version of the   following more general single-crossing condition \cite{Top98}: For  $\bar{\obs} > \obs$ 
 \beq (\reward_{a+1} - \reward_a )^\p B_{\bar{\obs}} \belief \leq 0 \implies (\reward_{a+1} - \reward_a )^\p B_{\obs} \belief \leq 0.
\label{eq:scr}
 \eeq This single crossing condition is ordinal, since  for any monotone function $\phi$,
it is equivalent to
$$ \phi( (\reward_{a+1} - \reward_a )^\p B_{\bar{\obs}} \belief ) \leq 0 \implies \phi( (\reward_{a+1} - \reward_a )^\p B_{\obs} \belief)  \leq 0.$$

\subsection{Structure of Public Belief Update}
We assume in  this section that the observation space and action space of the anticipatory agent
are $
\obspace = \{1,\ldots,\obsdim\}$, $\actionspace_2 = \{1,2\}$.
The purpose of this section is to show that even though
the public belief $\belief \in \Belief$ is continuum, there are only $\obsdim+1$ possible distinct action likelihood probability matrices. 

Specifically, define the following $\obsdim$ points in the one-dimensional simplex $\Belief$:
$$ \belief^*_\obs = \{\belief: (\reward_1 - \reward_2)^\p \oprob_\obs \tp^\p \belief = 0\}, \quad \obs = 1,\ldots, \obsdim $$

Note that $\belief_\obs^* = [1-\belief_\obs^*(2), \; \belief_\obs(2)]^\p$ depends on $\action_1,\physical$.

\begin{theorem} \label{thm:globalbelief} Under \ref{itemtp2}, \ref{supermod2},
it follows that \beq \belief_1^*(2) \leq \belief_2^*(2) \cdots \leq \belief_{\obsdim}^*(2) \label{eq:orderedthres} \eeq
Thus  the belief space
  $\Belief=[0,1]$ can be partitioned into at most $\obsdim+1$ non empty intervals denoted
  $\mathcal{P}_1,\ldots,\mathcal{P}_{\obsdim+1}$ where
  \beq \mathcal{P}_1 = [0, \belief_1^*(2)], \mathcal{P}_2 = (\belief_{1}^*(2), \belief_2^*(2)] , \ldots, \mathcal{P}_{\obs+1} = (\belief_{Y}^*(2),1]  \eeq
  On each such interval, the action likelihood $\oprobg$
(\ref{eq:actionprob})
  is a constant with respect to belief $\belief$. Specifically,  for fixed $\action_1,\physical$   
  \beq
\oprobg(s) = \begin{bmatrix} \sum_{i=0}^{l-1} \oprob_{1i}  &
  \sum_{i=l}^{\obsdim} \oprob_{1i} \\
  \sum_{i=0}^{l-1} \oprob_{1i} & \sum_{i=l}^{\obsdim} \oprob_{1i}
\end{bmatrix}, \quad \belief \in \mathcal{P}_l \label{eq:oprobg1}
\eeq
\end{theorem}

{\em Example}.
For $\obsdim=3$, the 4 possible  action likelihood
matrices $\oprobg$ are
\begin{equation} \label{eq:ex31} \begin{split}
R^1(\physical) = \begin{bmatrix} 0 & 1 \\ 0 & 1 \end{bmatrix},\;
R^2(\physical)  = \begin{bmatrix} B_{11}  & B_{12} + B_{13} \\ 
					B_{21} & B_{22}+B_{23} \end{bmatrix},\\
R^3(\physical) = \begin{bmatrix}  B_{11} + B_{12} & B_{13} \\ 
					   B_{21} + B_{22} & B_{23} \end{bmatrix}, \;
R^4(\physical) = \begin{bmatrix} 1 & 0 \\ 1 & 0 \end{bmatrix}	.				   
\end{split}
\end{equation}

Although  tangential to this paper,  agents deploying Protocol~1 can exhibit herding behavior. i.e., agents choose actions independent of their private observations;  see \cite{Cha04,Kri12}  for the distinction between herds and information cascades.

\subsection{Quickest Detection with Anticipatory  Agents  is non-trivial}  

In classical quickest change detection, the value function is always concave and the optimal stopping region is convex, see \cite{Kri16} for a partially observed Markov decision formulation and proof of this. The aim of this section is to show that due to the interaction of local and global decision makers,  quickest detection with anticipatory agents exhibits non-trivial behavior:  the value function is not necessarily concave and the stopping region is not necessarily a convex set.

Consider the value iteration algorithm (\ref{eq:vi}) which is used as a basis for mathematical induction to prove properties associated with
Bellman's equation (\ref{eq:bellman}).  Note that from (\ref{eq:vi}),
$\valueg_k(\belief,\physical)$ is positively homogeneous, that is, for any $\alpha > 0$, $\valueg_k(\alpha \belief,\physical) = \alpha \valueg_k(\belief,\physical)$.
So choosing $\alpha = \sigma(\belief,a)$  yields 
\begin{align} &\valueg_{k+1}(\belief,\physical) =
                \min\big\{\Cost(\belief,1) \label{eq:vinc}  \\ & + \int_{\physicalstatespace}   \sum_\action \sum_{l=1}^{\obsdim+1}\valueg_k(R_a^l(\physical)  \tp^\p \belief, \physical) I(\belief\in  \mathcal{P}_{l}) \pdf(\bar{\physical}|\physical) d\bar{\physical} , \Cost(\belief,2) \big\}
                                                             \nonumber
\end{align}
Recall $\Cost(\belief,1)$ and $\Cost(\belief,2)$ are linear in $\belief$.
However, it is clear from (\ref{eq:vinc}) that  if $\valueg_k(\belief,\physical)$ is assumed to be concave on $\Belief$, $\valueg_{k+1}(\belief,\physical)$ is not necessarily
concave on $\Belief$; since patching together convex functions on different intervals does not necessarily yield a convex function. The key point is that  the action  likelihoods $\oprobg$  (\ref{eq:aprob}) 
 are explicit and discontinuous functions of 
$\belief$. This results in a possibly
 non-concave value
function $V(\belief)$ making the stopping set  $\Stop$ non-convex.


\subsection{Numerical Example of Multi-threshold Quickest Detection Policy:  Change-Blindness} \label{sec:changeblind}
The non-concave value function in quickest detection with anticipatory agents leads to unusual multi-threshold behavior in the optimal policy, as we now illustrate.

\subsubsection{Setup} \label{sec:setupnumerical}
Consider  quickest detection  where the state of nature $\{\state_\dtime,\dtime\geq 0\}$  jumps according to transition matrix 
\beq \tp = \begin{bmatrix} 1 & 0 \\ 0.05 & 0.95 
\end{bmatrix}. \label{eq:tpval}\eeq
The global decision maker's delay and false alarm penalties are $d = 1.05, f = 3$; these specify the costs
(\ref{eq:cp1}), (\ref{eq:cp2}) in Bellman's equation (\ref{eq:bellman}).

The local anticipative  decision maker's reward matrix is
$$ (\reward_2(\state,\action_2),\state \in \{1,2\}, \action \in \{1,2\})
 = \begin{bmatrix}   5 & 4 \\ 6.5 & 9
\end{bmatrix}$$ Also its observation likelihood matrix is
$\oprob = \begin{bmatrix} 0.9 & 0.1 \\ 0.1 & 0.9 
\end{bmatrix}$.

\subsubsection{Nonconvex Stopping Time and Value Function}
\begin{figure}[h]
  \centering
  \begin{subfigure}{.45\textwidth}
    \includegraphics[scale=0.35]{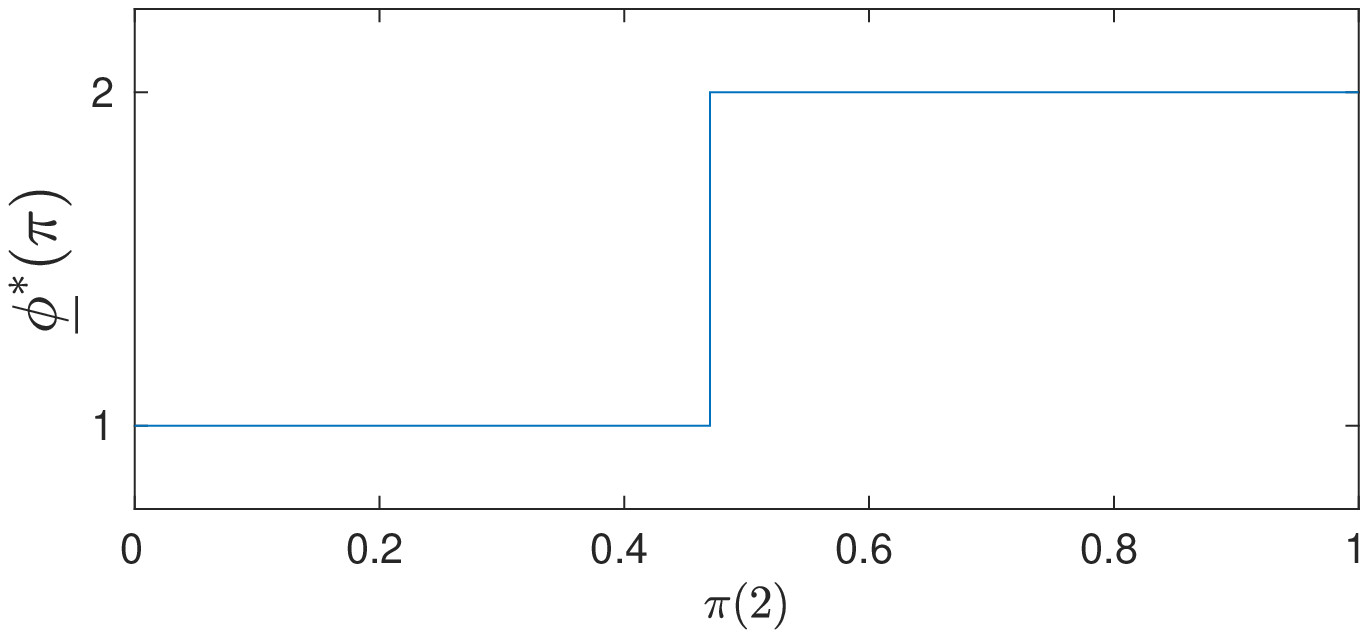}
    \caption{Optimal Policy}
  \end{subfigure}
  \begin{subfigure}{.45\textwidth}
\includegraphics[scale=0.35]{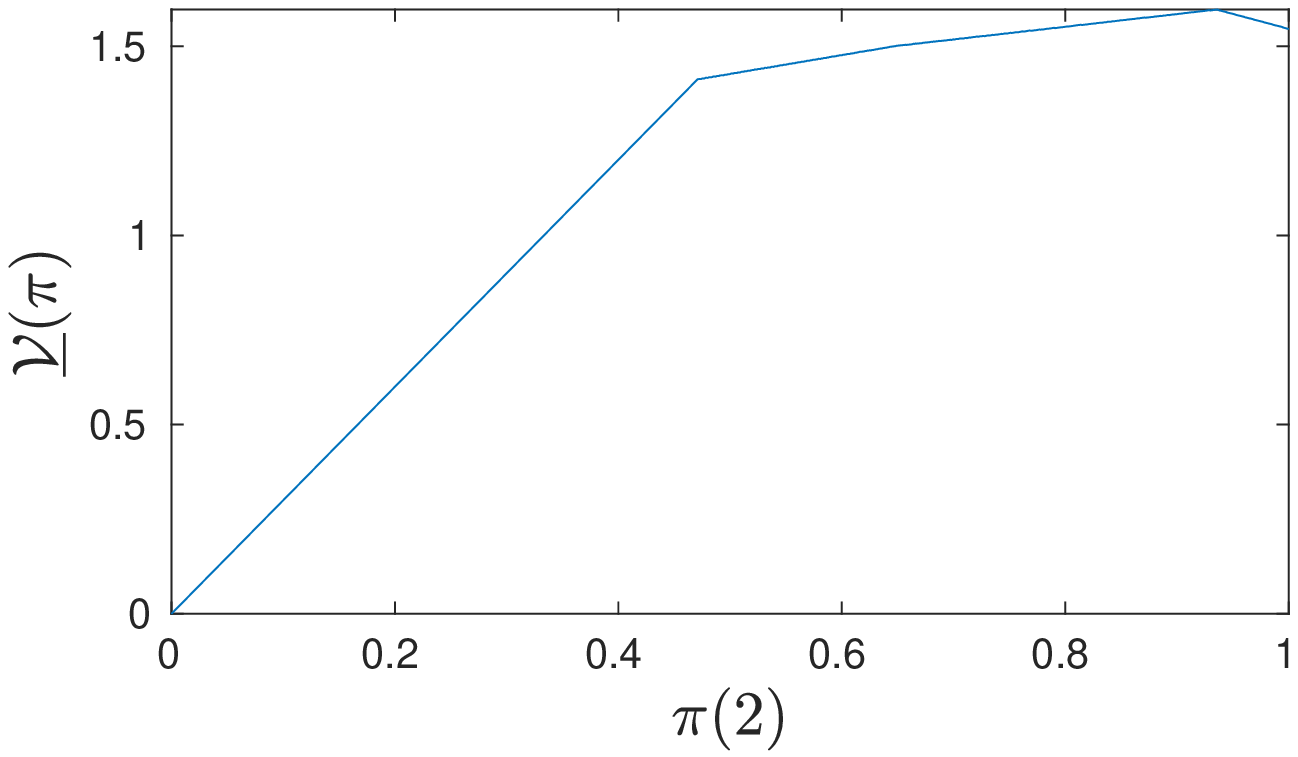}
 \caption{Value Function }
\end{subfigure}
\caption{Classical Quickest Detection. The optimal policy $\umu^*(\belief)$ has a threshold structure. So  the optimal stopping set $\Stop = \{\belief: \umu^*(\belief) = 2\}$ is convex. The
  value function $\uvalueg(\belief)$ is concave.}
\label{fig:classicalqd}
\end{figure}

\begin{figure}[h]
  \centering
  \begin{subfigure}{.45\textwidth}
  \includegraphics[scale=0.38]{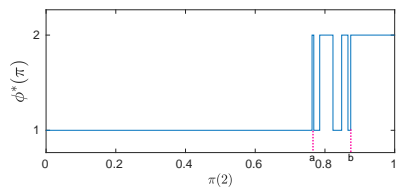}
\caption{Optimal Policy}
\end{subfigure}
\begin{subfigure}{.45\textwidth}
  \includegraphics[scale=0.38]{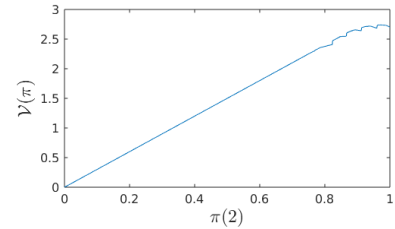}
   \caption{Value Function}
 \end{subfigure}
 \caption{Quickest Detection with Multiple  Agents. The optimal policy $\globalpolicy^*(\belief)$ has a multi-threshold structure implying that the optimal stopping set $\Stop = \{\belief: \globalpolicy^*(\belief) = 2\}$ is not  convex (comprises of disconnected regions). The global decision maker exhibits  change blindness. 
As the posterior probability  of change $\belief(1) = 1-\belief(2)$ increases from $b$ to $a$, the global decision maker declares there is no change in several regions.
The
   value function $\valueg(\belief)$  is not~concave.}
 \label{fig:anticipatoryqd}
\end{figure}

The local and global decision makers operate according to Protocol 1. Figure \ref{fig:classicalqd}  displays  the value function and optimal policy for classical quickest detection.
Figure \ref{fig:anticipatoryqd}  displays  the value function and optimal policy for  quickest detection with anticipatory agents. The policy and value function were obtained by running the value iteration algorithm
for 1000 iterations with $\Belief=[0,1]$ grid quantized uniformly  to 1000 values. 

For classical quickest detection, Figure \ref{fig:classicalqd} shows that, as expected,  the value function is concave and the optimal policy is a threshold. So the stopping region $\{\belief: \umu^*(\belief) = 1\}$ is the interval  $\belief(2) \in [0,0.466]$.

In contrast for quickest detection involving anticipatory agents, Figure \ref{fig:anticipatoryqd} shows   the value function is not concave. Also the optimal policy has an unusual  multi-threshold structure: if it is optimal to declare a change for a particular posterior probability, it may not be optimal to declare a change when the posterior probability of change is larger!  (Recall $1-\belief(2)$ is the posterior probability of change).
In this sense,  Figure \ref{fig:anticipatoryqd} depicts two forms of change-blindness. First, a  human global decision maker might choose to ignore the optimal policy $\globalpolicy^*(\belief)$ and simply use the classical quickest detection policy $\umu^*(\belief)$. A second, and  more interesting form  of change-blindness occurs when the human  global decision maker chooses the ``simple'' stopping set as $\belief(2) \in [0,a]$ and ignores the important  regions between $[a,b]$ where it is optimal to stop.

\subsection{Numerical Example.  Change Detection of Betting Strategy} \label{sec:spot}

  Spot fixing
  is a form of illegal match fixing where
  players  deliberately under-perform  in specific segments of a team sport. Identifying spot fixing  in cricket and soccer is  important with the advent of live betting. Sudden increases in the betting rate,  heavy underdog bets and wide swings in the quality of play  can prompt monitors to take a closer look at a match.
Quickest change detection of these  parameters based on monitoring real time betting is relevant for detecting illegal spot-fixing for example in T20 cricket; see also \cite{AJ19,FM19,QV13}.
Here we consider a highly simplified formulation where the aim is detect a sudden change in the intrinsic value (state of nature $\state_\dtime$) of the bet possibly due to spot fixing.

\subsubsection{Model}
Suppose each anticipatory betting agent $\dtime$ acts according to Sec.\ref{sec:bet} and makes decisions $\acta_\dtime\in  \{\bet,\nobet\}$, $\actb_\dtime
\in [0,\wealthmax]$.
The agents $\dtime=1,2,\ldots$  act sequentially according to Protocol 1. Each agent $\dtime$ has access to whether the previous agents placed bets, i.e., agent $\dtime$ knows the actions
$\{\acta_l\in \{\bet,\nobet\}, l=1,\ldots\dtime-1\}$.
The state of nature $\state_\dtime$ is the underlying value of the bet.
Each agent $\dtime$  obtains a noisy value $\obs_\dtime$ of $\state_\dtime$; this
determines
its  private belief  $\private_\dtime$ of $\state_\dtime$.
As in \eqref{rau2}, we assume that each agent is risk averse
and we choose its
risk averse parameter $\beta = \private_\dtime(1) $, i.e., the risk averse parameter of agent $n$ is its belief of the underlying value of the bet.
The current score in the game
 (physical state $\physical_\dtime$) also affects $\beta$; but for simplicity we omit this.

An analyst monitors the betting decisions $\{\acta_\dtime\}$. Due to privacy constraints, the amounts bet $\{\actb_\dtime\}$  are not known
to the analyst. How can the analyst detect a sudden change in the intrinsic value of the bet $\state_n$ indicating spot fixing?

We chose the anticipatory model parameters  $u_\Ali =10$, 
$g=15$, $\gardenreward=0$ (recall notation in \eqref{eq:polbet}).
The transition probability $\tp$ for the jump change and observation probabilities $\oprob$ are as Sec.\ref{sec:setupnumerical}. The quickest detection penalties are $d=1$, $f=10$. The system operates according to Protocol 1.

\subsubsection{Non-concave non-monotone Value Function}
\begin{figure} \centering
  \includegraphics[scale=0.37]{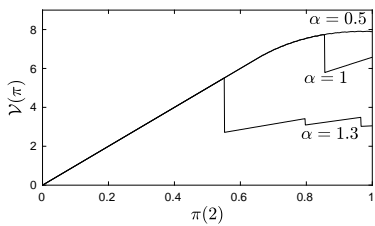}
  \caption{\blue{Non-concave value function for quickest change detection in betting strategy of anticipatory agents. The  parameters are specified in  Sec.\ref{sec:spot}.}}
  \label{fig:bet}
\end{figure}

Figure \ref{fig:bet} displays the quickest detection value function $\valueg(\belief)$
and optimal policy $\globalpolicy^*$  (\ref{eq:bellman}).
Unlike classical  quickest detection the value function is non-concave and not increasing, but the optimal policy (not shown) still has a  threshold structure.
Even this  simplistic example shows a rich variation of  the value function as $\alpha$ is adjusted: if $\alpha=0.5$, the $\valueg(\cdot)$ is concave;
if $\alpha=1.3$ is non-concave with multiple discontinues in $\valueg(\cdot)$.

 \subsubsection{Performance of Quickest Detectors}
\begin{figure}[h]
  \centering
  \begin{subfigure}{.23\textwidth}
  \includegraphics[scale=0.4]{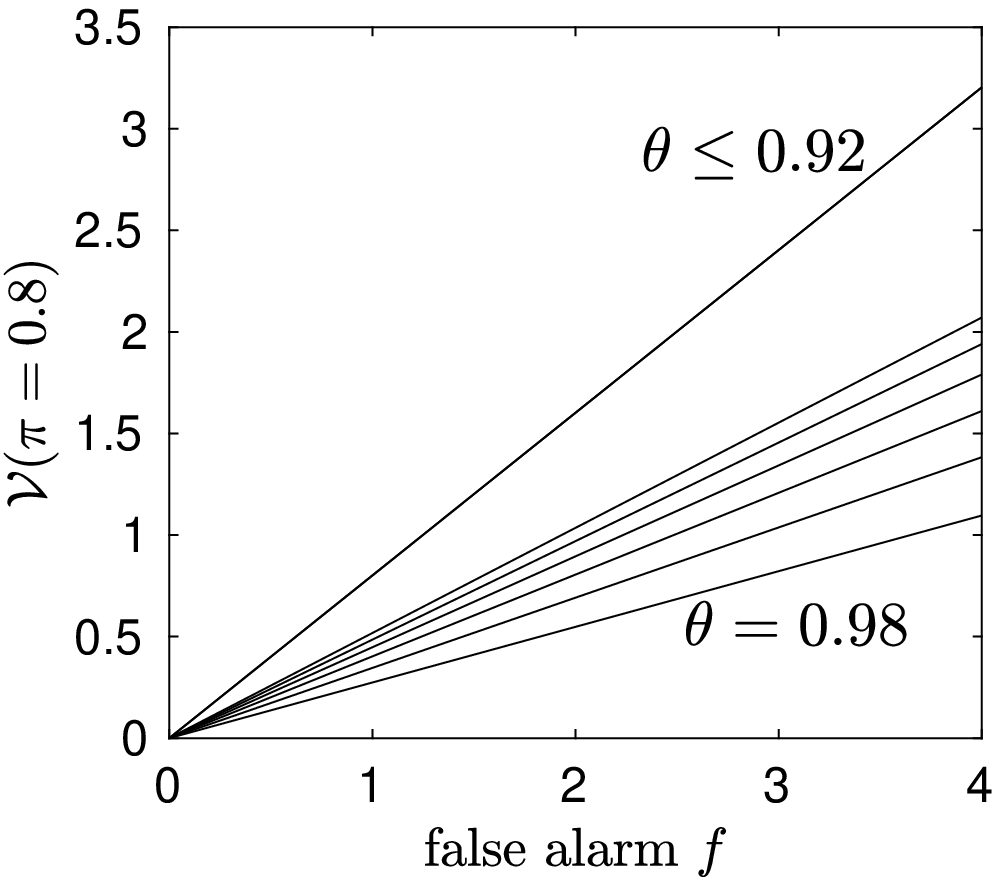}
\caption{Multi-agent}
\end{subfigure}
\begin{subfigure}{.23\textwidth}
  \includegraphics[scale=0.4]{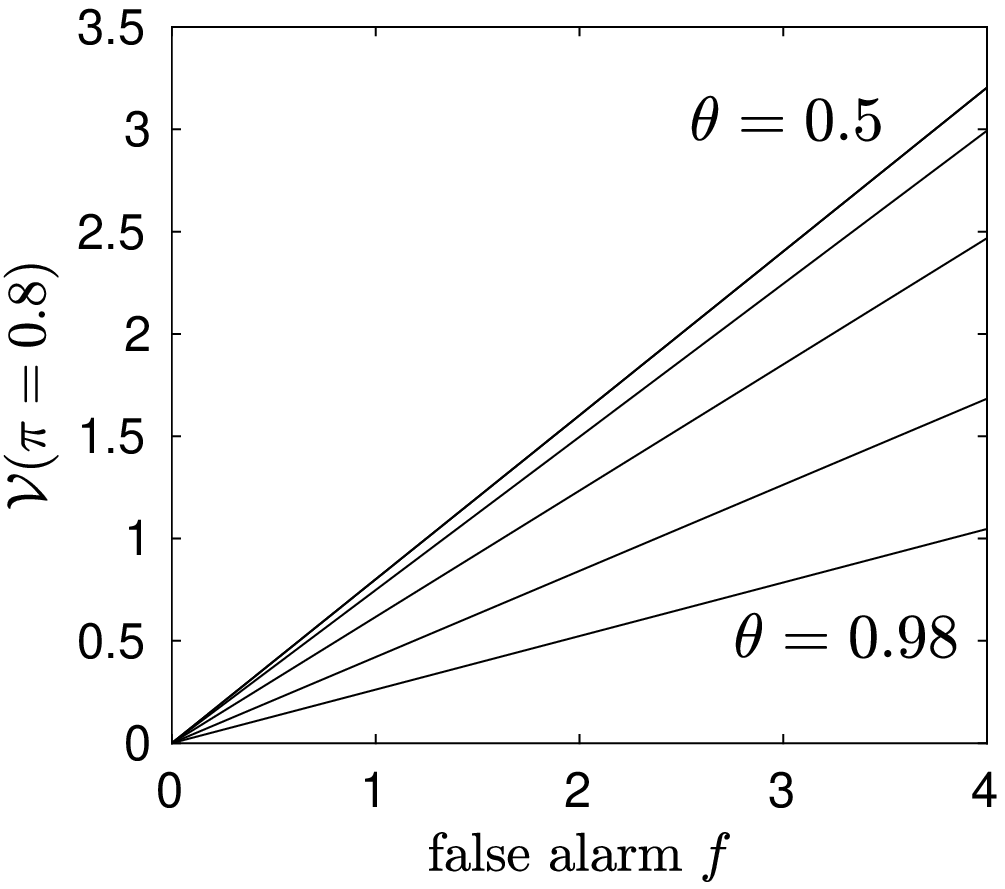}
  \caption{Classical}
 \end{subfigure} 
 \caption{\blue{Comparison of Optimal Expected Cost for Quickest Detection with Anticipatory Agents vs  Classical Quickest Detection}}
 \label{fig:performance2}
\end{figure}

  Figure \ref{fig:performance2} compares the performance of the  quickest detector with
  anticipatory    agents vs the classical quickest detector using the same parameters as above
  with $\alpha=1$. The observation probability matrix is  $\oprob=\begin{bmatrix} \oprobparam & 1 - \oprobparam \\  1 - \oprobparam & \oprobparam
\end{bmatrix}$ where  parameter $\oprobparam $ is varied. The delay penalty is fixed at $d=1$ while the false alarm $f\in [0.2,4]$.
The optimal expected  cost  $\valueg(\cdot)$ is obtained by solving Bellman's equation (\ref{eq:bellman}) by quantizing the beliefs to a grid. We chose $\belief = [0.2,\; 0.8]^\p$
in the plot
since the value function has a discontinuity just after $\belief(2) = 0.8$.


It is interesting to note that for quickest detection with multiple  agents, the optimal expected cost remains the same for $\oprobparam\leq 0.92$.
Another point to note is that the optimal cost is always larger than classical quickest detection. This is justified in Theorem \ref{thm:blackwell} below via Blackwell dominance.

\vspace{-0.7cm}

\subsection{Blackwell Dominance Implications for  Optimal Cost}
\label{sec:blackwell}
In this section  we  show that
quickest detection with anticipative agents (Protocol 1) results in a cumulative Kolmogorov Shiryaev cost $\globaltogo_{\globalpolicy^*}(\belief,\physical)$  (defined in (\ref{eq:ksd}) or equivalently (\ref{eq:beliefcost}))  that is  always larger  than that of classical quickest detection.
In Protocol 1,  agents have access to the public belief (which depends on local decisions of previous agents) instead of the 
actual observations. One expects that  this information loss results in less
efficient quickest time change detection compared to classical quickest detection.
Here we confirm this intuition.
The main idea involves  Blackwell dominance of observation measures. The result is useful because even though explicit computation of the optimal policy for the setup in Protocol 1  is difficult, we can lower bound the optimal achievable cost by that of classical quickest detection.

First define the optimal policy and cost in classical quickest change detection.
Similar to (\ref{eq:bellman}), the optimal policy $\umu^*(\belief)$ and cost $\uvalueg(\belief)$ incurred in classical quickest detection, satisfy  the following stochastic  dynamic programming equation:
\begin{align} \label{eq:dp_algc}
\umu^*(\belief)&= \arg\min_{\globalaction \in \globalactionspace} \uQ(\belief,\globalaction) , \;  \uvalueg(\belief) = \min_{\globalaction \in \globalactionspace} \uQ(\belief,\globalaction),\\
 \text{ where } &  \uQ(\belief,2) =  C(\belief,2)
+ \sum_{y \in \obspace}  \uvalueg\left( T(\belief ,y) \right) \filterd(\belief,y),\nonumber\\
& \uQ(\belief,1) =   C(\belief,1) , \qquad
  \uglobaltogo_{\mu^*}(\belief) =  \uvalueg(\belief).   \nonumber
\end{align} 
Here  $T(\belief,y)$ is the  Bayesian filter update defined in (\ref{eq:privateupdate}) and
$  \uglobaltogo_{\mu^*}(\belief)$ is the cumulative cost of the optimal policy starting with initial belief $\belief$.
Note that unlike  Protocol 1, in  classical
quickest detection, there is no public belief  update 
(\ref{eq:pubupdate}) or interaction between the public and private beliefs.

The following theorem says that for any  initial belief $\belief$,
  the optimal detection policy with anticipative agents acting sequentially (Protocol 1)  incurs a higher cumulative cost  than that of classical quickest detection.

\begin{theorem}\label{thm:blackwell} Consider the  quickest change detection problem involving anticipatory  agents described in Protocol 1 and    associated value function $\valueg(\belief,\physical)$ in (\ref{eq:bellman}). Consider also the classical quickest detection
problem with value function $\uvalueg(\belief)$ in (\ref{eq:dp_algc}).
 Then for any initial
 belief  $\belief \in \Belief$, the optimal cost incurred by classical quickest detection
is smaller than that of quickest detection with anticipatory agents. That is, $\uvalueg(\belief) \leq \valueg(\belief,\physical)$ for all $\belief \in \Belief, \physical \in \physicalstatespace$.
\end{theorem}

The proof is in the supplementary document.
 
 The intuition behind the proof is as follows.
 From (\ref{eq:oprobgcompute}) 
\begin{equation} \begin{split} \oprobg_{\state,\action}(\physical)  &= \int_{\obspace}  \oprob_{\state,\obs}  \Mb_{\obs,\action,\physical} d\obs , \\
\text{ where } \Mb_{\obs,\action,\physical} &\ole  I(\polpT(\physical,\action_{\dtime,1}) = \action)
\end{split} \label{eq:aprob}
\end{equation}
where 
 $\oprob$ and $\Mb$ are stochastic kernels.  Thus 
observation $\obs$ with conditional distribution specified by $\oprob$ is 
said to be more  informative than (Blackwell dominates) observation $\action$ with conditional distribution $\oprobg$,
see \cite{Kri16}. The main idea in the proof is that under the assumptions of Theorem~\ref{thm:blackwell},
the value function $\uvalueg(\belief)$ is concave for $\belief \in \I$. Then the result is established
 using Jensen's inequality together
with Blackwell dominance  on  Bellman's equation (\ref{eq:bellman}).
 
A useful consequence of  Theorem \ref{thm:blackwell} is that performance analysis of standard quickest detection \cite{TV05} 
applies  as a lower bound for  quickest detection with anticipatory agents.

\vspace{-0.4cm}

\section{Discussion}
This paper is an early step in addressing  sequential detection problems with behavioral economics constraints. Although both signal processing and behavioral economics are mature areas,
 insights gained by construction of generative anticipatory  models, estimation algorithms, along with careful analysis  is crucial in designing human-sensor cyber-physical systems.

They main results of the paper are: \\
1.  Formulation of the  two stage decision making model of \cite{CL01} for individual decision makers involving the anticipatory state.  The key idea is that the anticipatory  state involves the probabilities of future actions thereby leading to time inconsistency in decision making.\\
2.  Characterizing the structure of the subgame Nash equilibrium as a bang-bang controller in the first time stage, and a threshold policy at the second time stage (Theorem~\ref{thm:nashstructure}).
  The bang-bang structure justifies the observation in \cite{CL01} that agents with anticipatory emotions may choose to  avoid information. \\
3.  Formulation of the multi-agent quickest detection problem where the anticipatory agents interact with a global decision maker. We gave several examples to motivate this problem including change detection in social media accommodation, detecting spot fixing in sports and  team-situation awareness. \\
4.  Structural characterization of the unusual structure of the optimal change detection policy (compared to classical quickest detection).  Sec.\ref{sec:structure2} characterized the structure of the Bayesian belief updates and achievable cost of the quickest detector without brute force computations. We derived  important  structural properties of the Bayesian updates of the local and global decision makers (Theorem \ref{thm:monotone} and Theorem~\ref{thm:globalbelief}), constructed a lower bound for the optimal cost incurred using Blackwell dominance (Theorem \ref{thm:blackwell}), and presented numerical examples of the unusual structure of the optimal quickest change policy (non-convex stopping region). The multi-threshold change detection  policy was interpreted as  {\em change blindness}, namely people fail to detect surprisingly
large changes to scenes.

In future work we will generalize  the anticipatory model using the subjective belief multi-horizon formulation of \cite{BPP17}.  It is also worthwhile conducting a  performance analysis of a multi-threshold detector; see  \cite{TV05}  for performance analysis  involving a single threshold detector. \blue{An important open question is: based on a dataset of actions of an agent, how to  identify anticipatory behavior and if so,  how to estimate the utility function of an anticipatory agent
  (inverse reinforcement learning)?
For   myopic Bayesian utility maximization, \cite{CD15} give  necessary and sufficient conditions for identifying
optimal behavior; the utility functions then are feasible points of a set of convex constraints.  In \cite{HKP20} we have used such methods to analyze user engagement in massive YouTube datasets.}


\newpage

{\large Supplementary Document}

\begin{abstract} The main paper gave a complete description of anticipatory decision making and quickest change detection with anticipatory decision makers. This supplementary document contains a detailed example of anticipatory decision making in terms of  a social media accommodation example. Then proofs of theorems stated in the main paper are given.

\end{abstract}

\section{Tutorial Example. Social Media based Accommodation Choice} \label{sec:airbnb} We now discuss an anticipatory decision making example involving choosing accommodation using a social-media based online agency such as Airbnb. The example is a slight generalization of   \cite{CL01} since  the rewards are parametrized by a Bayesian posterior  $\private$ (the motivation for this in terms of quickest detection is discussed in the paper).

\subsection{Single Anticipatory Agent} 
Suppose an anticipatory agent chooses between  a vacation either at a previous known accommodation $(\known)$, or
a new accommodation ($\new$). The agent has initial wealth of $\wealth_0$.

\subsubsection{Model} The reviews of accommodation $\new$ posted at an online reputation website at times $k=1,2$ are  $\weather_ k  \in 
 \{G \text{ (good)},B \text{ (bad)}\}$.
 The physical states  $\physical_1$ and $\physical_2$ denote the probability of good review  of
 $\new$ at times 1 and 2. For simplicity, assume $\physical_2$ is uniformly distributed in $[0,1]$.

 Regarding the actions,
at the first stage the agent  chooses action
$\action_1 \in [0,1]$ which denotes making a  non-refundable deposit  $2000 \,\action_1$  for booking $\known$.
At the second stage, the agent makes the final choice of which accommodation to stay in, i.e., $\action_2 \in 
\actionspace_2= \{\known,\new\}$. 

Next we model the anticipatory emotions of the agent.
Similar to \cite{CL01}, we choose the anticipatory reward to reflect beliefs about pleasure that will be derived in staying respectively at venues $\known$ and $\new$.
We choose the psychological (anticipatory) state $\psych_1$ at time 1 as  the conditional probabilities
(see (\ref{eq:psych}))
\beq \begin{split}  \psych_1 = \max\{ & 6000\, \pdf(\action_2=\new, \weather_2=G| \action_1,\policy_2) ,  \\  & 4000\, \pdf(\action_2 = \known| \action_1,\policy_2)\}
\end{split}  \label{eq:psych_example} \eeq
So the anticipatory pleasure increases with the agent's certainty that an outcome will occur.
Also (\ref{eq:psych_example}) specifies that the anticipatory pleasure is higher for $\new$ (since it scaled by 6000) compared to $\known$  providing that $\weather_2$ is good.

We now construct the rewards $\reward_1, \reward_2$ defined in (\ref{eq:cl_cost}).
\begin{compactenum} \item
Assume
each accommodation costs 2000 units.
\item After making a deposit of $2000 \action_1$ for $\known$,
  if $\new$ is chosen, then the deposit of $2000\action_1$ is lost.
  \item The benefit accrued by staying in $\new$
when rating is $\physical_2$ is $6000 \physical_2 \thbelief$; the reward for choosing  $\known$ is 4000.
Here\footnote{\label{foot:private}We assume $\private = [\private(1),\private(2)]^\p$ is a 2-dimensional probability vector, i.e., $\beliefdim=2$ in (\ref{eq:thbelief}).  For rotational convenience, we refer to $\private(2)$ as $\private$.} $\thbelief\in [0,1]$ is the posterior  probability that accommodation $\new$ is suitable given the most recent review of $\new $.
\item Finally,
  $\beta>0$ denotes  the importance of anticipatory reward relative to the reward of the vacation (see \ref{beta}).
\end{compactenum}
Based on  the above description, the  rewards are
\begin{equation*}
  \begin{split}
&  \reward_1 = \beta \psych_1,  \quad \beta > 0  \\
& \rewardp(\physical_2,\action_2=\new, \action_1) = 6000 \physical_2 \thbelief+ \wealth_0 - 2000 (1+\action_1) , \\
& \rewardp(\physical_2,\action_2=\known,\action_1) = 4000 + \wealth_0 - 2000
\end{split}
\end{equation*}




\subsubsection{Structural Result for Nash equilibrium} For the above example, we can verify Assumptions   \ref{actionspace}-\ref{convexcdf} hold and therefore Theorem \ref{thm:nashstructure} holds. Specifically, \ref{actionspace} holds by formulation; \ref{convexr2}  holds trivially since $\rewardp$ is linear in $\action_1$; \ref{supermod} holds since $\rewardp(\physical_2,\action_2=\known,\action_1)$ is  independent  of $\physical_2$; \ref{implicit} and  \ref{convexthreshold} hold trivially since $\diffreward$ is linear in $\physical_2$ and $\action_1$; \ref{beta} holds by construction since it is  easily shown that for optimal policy  $\pol_2^*$,
$z_1 = 4000\, \pdf(\action_2 = \known| \action_1,\policy_2^*)$. Finally,  \ref{convexcdf} holds since $\pdf(\physical_2)$ is the  uniform density. 

Therefore from Theorem \ref{thm:nashstructure} it follows that $\pol_2^*$ has a threshold structure (\ref{eq:pol2threshold}), and $\pol_1^*$ has a   bang-bang structure (\ref{eq:bangbang}).

Therefore the interpretation of deliberate avoidance of information discussed below
Theorem \ref{thm:nashstructure} holds.
Specifically,  due to the bang-bang structure of (\ref{eq:bangbang}),  the agent makes a full deposit $\action_1 = 1$ if $\beta > \beta^*$ for the accommodation $\known$.  Yet this full non-refundable deposit does not guarantee that the agent will choose $\action_2 = \known$ since if $\physical_2 > 
\physical_2^*(\action_1) $, then  the agent will choose $\action_2 =\new $. Thus the agent might deliberately choose not to observe the state $\physical_2$ in order not to lose the deposit paid at time 1 to secure $\known$.

\subsubsection{Explicit Evaluation of Nash equilibrium}
Given the  simple structure above, we can go beyond Theorem \ref{thm:nashstructure} and 
solve explicitly for the subgame Nash equilibrium specified by (\ref{eq:period2}), (\ref{eq:period1}). The computations below are similar to \cite{CL01}.

 From the extended Bellman equation
 (\ref{eq:period2}), $\policy_2^*(\physical_2,\action_1)$ has threshold structure 
\beq \label{eq:pol2*}
 \polp2(\physical_2,\action_1) = \argmax_{\action_2} \rewardp(\physical_2,\action_2,\action_1)= \begin{cases} \new &  \text{ if } \physical_2 \geq \frac{2 + \action_1}{3\thbelief} \\
      \known   & \physical_2 < \frac{2+\action_1}{3\thbelief} 
    \end{cases} \eeq
    with associated value function
    \begin{multline*}\utility_2(\physical_2,\policy_2^*) = 
\max_{\action_2}  \{ \rewardp(\physical_2,\action_2,\action_1) \} \\  =
\begin{cases}  6000 \physical_2 \thbelief + \wealth_0 - 2000 (1 + \action_1)   & \physical_2 \geq \frac{2+ \action_1}{3\thbelief} \\
4000 + \wealth_0 -   2000 & \physical_2 < \frac{2+\action_1}{3\thbelief}
\end{cases} 
\end{multline*}

In order to determine  the policy $\policy_1^*$ and value function $\utility_1$, let us first compute the psychological state $\psych_1$ in (\ref{eq:psych_example}) under $\policy_2^*$. 
Since $\physical_2 $ is uniformly distributed in $[0,1]$,  clearly 
\begin{equation} \begin{split}
    \pdf(\action_2&=\known|\action_1,\policy_2^*) = \prob(\{\physical_2: \policy_2^*(\physical_2)= \known\}| \action_1)\\
    &=
    \int_\physicalstatespace \pdf(\physical_2|\physical_1) I(\physical_2:\policy_2^*(\physical_2=\known))\, d\physical_2 \\
    &= \int_\physicalstatespace \pdf(\physical_2)\, I(\physical_2\in [0,\frac{2+\action_1}{3\thbelief}]) d\physical_2 \\ &=
    \min\{\frac{2+\action_1}{3\thbelief},1\}
\end{split}
\end{equation}
\begin{equation*} \begin{split}
 &   \pdf(\action_2=\new,\weather_2=G|\physical_2,\action_1,\policy_2^*) \\ &=
    \pdf(\weather_2=G) \pdf(\action_2=\new|\action_1,\physical_2,\policy_2^*) = \physical_2\, I(\physical_2 \in [\frac{2+\action_1}{3\thbelief},1]) \\
  \end{split}
\end{equation*}
Therefore,
\begin{equation} \begin{split}
    &  \pdf(\action_2=\new,\weather_2=G\,|\action_1,\policy_2^*) \\ & = \int_{\physicalstatespace}  \physical_2\, I(\physical_2 \in [\frac{2+\action_1}{3\thbelief},1])  \pdf(\physical_2) d\physical_2 = \max\{\int_{\frac{2+\action_1}{3\thbelief}}^1 \physical_2 d\physical_2,0\}\\
    &=
   \max\{ \frac{9 \thbelief^2 - 4 -  4 \action_1 - \action_1^2}{18 \,\thbelief^2},0\}
  \end{split}
\end{equation}
Then using notation (\ref{eq:cond}) and  (\ref{eq:psych_example}),  the psychological state is 
\begin{multline}
\psych_1=
\max\{ 4000\,\pdf(\action_2=\known|\action_1,\policy_2^*),  \\ 6000 \pdf(\action_2=\new,\weather_2=G\,|\action_1,\policy_2^*)\} \\ =
4000 \, \pdf(\action_2=\known|\action_1,\policy_2^*)
\end{multline}
\blue{The last equality is verified  since comparing the two terms involves a scalar quadratic inequality in $\thbelief  \in [0,1]$.}

Then substituting  $\utility_2$ computed in  (\ref{eq:pol2*}) into  (\ref{eq:period1}) yields
$$ \utility_1(\physical_1) =  \max_{\action_1\in \actionspace_1} \{ \beta \psych_1 +   \int_0^1  \utility_2(\physical_2) d\physical_2 \} $$
It is easily verified that the expression within $\{ \cdot \}$  is convex in $\action_1$. Since $\actionspace_1=[0,1]$ is convex, the  maximum is achieved at an extreme point $\action_1=0$ or $\action_1 =1 $.
Thus the optimal policy $\policy_1^*$ at time 1 is a $(\beta,\private)$ dependent  bang-bang policy:
\beq  \polp1 = \begin{cases} 1 \text{ (full deposit) } & \text{ if } \beta > 1-3\private + \frac{9 \private^2}{4} \\ 0 \text{ (no deposit) }  & \text{ if } \beta \leq 1-3\private + \frac{9 \private^2}{4}
\end{cases} \label{eq:bangbangex}
\eeq
Recall $\private$ is a Bayesian parameter (see footnote \ref{foot:private}) that is defined as the private belief in Sec.\ref{sec:qdnew} in the context of quickest detection, and $\beta >0$ is a scaling constant \ref{beta}.

\subsection{Quickest Change Detection}
Here we comment on how the above example extends to social media based decision making such as  media based
accommodation systems.   Individual anticipatory agents make local  decisions sequentially whether to rent a  property; these decisions are affected by the reviews (decisions) of previous agents.
The  global decision maker (e.g. Airbnb) monitors these local decisions. How can the global decision maker  detect if there is a sudden change in the demand for a specific  accommodation due to the presence of a new competitor? Alternatively, how can the global decision maker detect a sudden change in the quality of an accommodation?

\begin{compactenum}
  \item The underlying state of nature $\state_\dtime$ is the intrinsic value of  the accommodation $\known$;  i.e., the ground truth. The value of $\state_\dtime$  depends on the cost of $\known$ and cost of  competitors.
    \item $\obs_\dtime$ is an extrinsic measurement made by the agent regarding  $\state_\dtime$.
    $\obs_\dtime$  is an interpretation of a competitor,  a recent photo/review  comparing $\known$ with a competitor, etc.
    \item
      $\public_{\dtime-1}$: This is the histogram available to the agent from the rating site comparing $\known$ with a competitor and so is reflective of $\pdf(\state_{\dtime-1}|\action_1,\ldots,\action_{\dtime-1})$.
    \item Recall $\physical_\dtime$ denotes the probability of good  reviews of~$\new$. As discussed in the main paper, by allowing the transition kernel of $\physical_\dtime$ to depend on the ground truth
      $\state_\dtime$, we allow for the reviews to change with changing ground truth.
      \item The agent updates its belief $\private_\dtime$ based on this information.  The Bayesian update  is a useful idealization\footnote{Theorem \ref{thm:monotone}   shows that if the  costs and observation probabilities to satisfy reasonable conditions, the decisions made by agents are {\em ordinal} functions of their private observations and {\em monotone}
  in the prior information. Thus  the Bayesian update  follows simple intuitive rules and is  a useful idealization.} of the agent's behavior; see Theorem \ref{thm:monotone} below. 
\item The agent then makes decisions  $\action_{\dtime,1}, \action_{\dtime,2}$  according to Protocol 1.
\end{compactenum}
Given the sequence of decisions $\{\action_\dtime\}$, the
 global decision maker (e.g. Airbnb)  wishes to detect if there is a sudden 
 appearance of competition or sudden change in quality of the accommodation.

\section{Human-Sensor System}
Quickest change detection
involves decision making in a partially observed Bayesian  setting. In the context of this paper, there are two interpretations.
\begin{compactenum} \item
In human-sensor interface systems,
each anticipatory agent is equipped with a sensing/computing device. The sensing device  observes the state of nature (Markov chain) in noise. The computing device evaluates   the posterior and provides the agent with these probabilities. The agent (human)  then makes  anticipatory decisions $\action_1,\action_2$  as detailed in Sec.\ref{sec:structure}.
\item The second interpretation is as follows:
 If the anticipative agent $ \dtime $  makes recommendations that are monotone and ordinal in the observations and monotone in the prior, then they mimic
the Bayesian social learning model.  Even if the agent does not exactly follow a Bayesian social learning model,  its monotone ordinal behavior implies that such a Bayesian model is  a useful idealization. 
Humans typically make {\em monotone} decisions - the more favorable the private  observation,
the higher the recommendation. Humans   make {\em ordinal} decisions.
\end{compactenum}

\section{Proofs of Theorems}

\subsection{Proof of Theorem \ref{thm:nashstructure}}

{\bf Remark}.
 {\em Relaxing Assumption  \ref{implicit}.} Instead of 
\ref{implicit}, the following weaker condition based on the classic implicit function theorem   \cite{Apo74}
  can be used. Assume  (i) $\diffreward(\physical_2,\action_1) $
has continuous first partial derivatives; (ii)
$\diffreward(\physical_2^*,\action_1) = 0 \implies \partial \diffreward(\physical_2^*,\action_1) /\partial \physical_2 \neq 0$. Then by the implicit  function theorem, the solution $\physical_2^*(\action_1)$ is continuously differentiable on an open subset of $(0,1)$. Assume that this subset is convex.

   \begin{proof} For convenience we omit parameter $\thbelief$ in the notation. 
     
  {\em Statement 1}.
  By \ref{supermod},
  $\rewardp$ is supermodular in $(\physical_2,\action_2)$. Thus by Topkis theorem \cite{Top98},  $\pol_2^*(\physical_2,\action_1) $ is non-decreasing in $\physical_2$ for fixed $\action_1$.
So either  $\pol_2^*(\physical_2,\action_1) $  is constant wrt $\physical_2$ (in which case the theorem holds trivially);  or for each $\action_1$ there
exists an indifference state  $\physical_2^* \in [0,1]$ such that $\pol_2^*(\physical_2,\action_1)$ switches from 1 to 2 as $\physical_2$ increases.
Clearly the indifference set   $\{\physical_2,\action_1: \diffreward(\physical_2,\action_1) = 0\}$ determines where  $\pol_2^*(\physical_2,\action_1) $ switches from 1 to 2. 
By \ref{implicit}, a solution $\physical_2^*(\action_1)$ exists to $\diffreward(\physical_2,\action_1) = 0$ for $\action_1\in (0,1)$.
Hence, $\pol_2^*$ has the threshold structure (\ref{eq:pol2threshold}).

  {\em Statement 2}.
 Proving statement~2, is equivalent to  showing convexity in $\action_1$ of the solution $\physical_2^*(\action_1)$
of 
the algebraic equation  $\diffreward(\physical_2,\action_1) = 0 $. By \ref{implicit},  $\physical_2^*(\action_1)$ is continuously
differentiable in $\action_1$.
 It is  verified by elementary calculus that
  $$ \frac{\partial^2 \physical_2^*}{\partial \action_1^2} = \frac{1}{(\partial \diffreward/\partial \physical_2)^2}\,\big[\frac{\partial \diffreward}{\partial \action_1} \frac{\partial^2 \diffreward}{\partial \physical_2 \action_1} - \frac{\partial \diffreward}{\partial \physical_2} \frac{\partial^2\diffreward}{\partial \action_1^2}\big]$$
  So $\physical_2^*(\action_1)$   is convex in $\action_1$ iff  \ref{convexthreshold} holds; see \cite{BT66} for a more general multidimensional result.

{\em Statement 3(a)}.  From \ref{beta}, the psychological state is 
  \begin{equation*} \begin{split}
   &   \psych_1  = \max_{\action \in \actionspace_2}\{ \int_\physicalstatespace I(\physical_2: \pol_2^*(\physical_2,\action_1) = \action)  \,
   \ant(\action,\physical_2)\,\pdf(\physical_2|\physical_1)\, d\physical_2\} \\
   & \quad = \max \{ \int_0^{\physical_2^*(\action_1)}  \ant(1,\physical_2)\,\pdf(\physical_2|\physical_1)\, d\physical_2, \\
   & \qquad \qquad \qquad
   \int_{\physical_2^*(\action_1)}^1  \ant(2,\physical_2)\,\pdf(\physical_2|\physical_1)\, d\physical_2 \}
      \\ & \quad =
      \max\big\{ \cdf_1( \physical_2^*(\action_1)) - \cdf_1(0),
      \cdf_2(1) - \cdf_2( \physical_2^*(\action_1)) \big\}\\
    &    \text{ where } 
  \cdf_\action( y) = 
 \int_0^{y} \ant(\action,\physical_2)\, \pdf(\physical_2|\physical_1)\, d\physical_2
    \end{split}
  \end{equation*}
 By  \ref{convexcdf}, $\cdf_1(y) $ and $-\cdf_2(y)$ are increasing convex functions of $y$.  Since $\physical_2^*(\action_1)$ is convex in $\action_1$ (by Statement 2), the composition
 functions $\cdf_1(\physical_2^*(\action_1)) $ and $-\cdf_2(\physical_2^*(\action_1))$
 are convex.
 Thus $\reward_1 = \beta \psych_1$ being the max of  two convex functions is convex in $\action_1$.
 This together with  \ref{convexr2} implies that the reward-to-go $\utilitytogo_1(\physical_1,\action_1,\pol_2^*)$ is convex in $\action_1$.

 {\em Statement 3(b)}.
Finally a convex function on a convex set (recall $\action_1 \in \actionspace_1= [ 0,1]$) achieves its global maximum  at an extreme point \cite[Theorem 3, pp.181]{Lue84}. Hence the bang-bang structure (\ref{eq:bangbang}) holds for $\pol_1^*$.
 \end{proof}

\subsection{Proof of Theorem \ref{thm:monotone}}

\begin{proof}
The proof uses MLR stochastic dominance (defined in footnote \ref{footnotemlr}) and the following  single crossing
condition:

\begin{definition*}[Single Crossing  \cite{Ami05}] \label{def:scc}
$g:\Y\times \A\rightarrow \reals $ satisfies a single crossing condition in  $(y,a)$  if $g(y,a) - g(y,\bar{a}) \geq 0$ implies $g(\bar{y},a)- g(\bar{y},\bar{a}) \geq 0$ for $\bar{a}>a$ and $\bar{y} > y$.
Then $a^*(y) = \argmin_{a} g(y,a)$ is increasing in $y$. \qed
\end{definition*}

By \ref{itemtp2}  it  follows  that \cite{Kri16} the Bayesian update satisfies
$$ \frac{\oprob_\obs \tp^\p  \belief}{\ones^\p \oprob_\obs \tp^\p  \belief}  \lr  \frac{\oprob_{\bar{\obs}} \tp^\p \belief}{\ones^\p \oprob_{\bar{\obs}} \tp^\p\belief}, \quad \bar{\obs} > \obs $$
where $\lr$ is the MLR stochastic order. (Indeed, the MLR order is closed under conditional expectation.)
By supermodularity \ref{supermod2}
$\reward_{a+1} - \reward_a $ is a vector with increasing elements.
Therefore 
$$(\reward_{a+1} - \reward_a )^\p \frac{\oprob_\obs \tp^\p \belief}{\ones^\p \oprob_\obs \tp^\p \belief}  \leq (\reward_{a+1} - \reward_a )^\p\frac{\oprob_{\bar{\obs}} \tp^\p \belief}{\ones^\p \oprob_{\bar{\obs}} \belief} $$
Since the denominator is non-negative, it follows that 
$ (\reward_{a+1} - \reward_a )^\p \oprob_{\bar{\obs}} \belief \leq 0 \implies (\reward_{a+1} - \reward_a )^\p \oprob_{\obs} \belief \leq 0$.
This implies that $ \reward_a ^\p \oprob_{\obs} \belief$  satisfies a single crossing condition in $(\obs,a)$.
Therefore $a_n(\belief,\obs) = \argmax_a \reward_a ^\p \oprob_{\obs} \belief$ is increasing in $\obs$ for any belief~$\belief$.
\end{proof}

\subsection{Proof of Theorem \ref{thm:globalbelief}}

\begin{proof}  The single crossing property~(\ref{eq:scr}) implies
$$
\{\belief: (\reward_1-\reward_2)^\p \oprob_\obs \tp^\p \belief \leq 0\} \subseteq
\{\belief: (\reward_1-\reward_2)^\p \oprob_{\bar{\obs}} \tp^\p \belief \leq 0\} $$
for $\obs < \bar{\obs}$. This implies (\ref{eq:orderedthres}).
From (\ref{eq:oprobgcompute}) we can write
\begin{equation} \begin{split} \oprobg_{\state,\action}(\physical)  &= \sum_{\obspace}  \oprob_{\state,\obs}  \Mb_{\obs,\action,\physical} , \\
\text{ where } \Mb_{\obs,\action,\physical} &\ole  I(\polpT(\physical,\action_{\dtime,1}) = \action)
\end{split} \label{eq:aprob}
\end{equation}
where 
 $\oprob$ and $\Mb$ are stochastic matrices. This yields (\ref{eq:oprobg1}).

\end{proof}

\subsection{Proof of Theorem \ref{thm:blackwell}} 

\begin{proof}

It is well known \cite{Kri16} (and straightforwardly demonstrated  by induction) that the value function  $\uV_k(\belief)$ for classical quickest detection  is concave over $\Belief$ for any $k$.
 We then use the Blackwell dominance condition (\ref{eq:aprob}). The  public belief update (\ref{eq:pubupdate}) can be expressed in terms of the private belief update  
 (\ref{eq:privateupdate}) as
\begin{multline}
 \filterg(\belief,\action,\physical ) =   \sum_{\obs  \in \obspace} \filter(\belief,\obs) \frac{\filterd(\belief,y)}{\filterdg(\belief,\action,\physical)}\, \Mb_{\obs,\action,\physical}  \\
 \text{ and } \filterdg(\belief,\action,\physical) = \sum_{\obs  \in \obspace} \filterd(\belief,y) \Mb_{\obs,\action,\physical}
\end{multline}
Note that  $\frac{\filterd(\belief,y)}{\filterdg(\belief,\action,\physical)} \Mb_{\obs,\action,\physical} $ is a probability measure wrt $y$.
Since  $\uvalueg_k(\cdot)$ is concave for $\belief \in \I$, using Jensen's inequality it follows that
\begin{align*}
\uvalueg_k(\filterg(\belief,\action,\physical) ) & = \uvalueg_k \left(\sum_{\obs\in \obspace} \filter(\belief,\obs) \frac{\filterd(\belief,\obs)}{\filterdg(\belief,\action,\physical)} \Mb_{\obs,\action,\physical} \right) \\
                            &\geq \sum_{\obs \in \obspace}  \uvalueg_k (T(\belief,y)) \frac{\filterd(\belief,\obs)}{\filterdg(\belief,\action,\physical)}  \Mb_{\obs,\action,\physical}
\end{align*}
Therefore for each $\bar{\physical} \in \physicalstatespace$,
$$  \sum_{a}  \uvalueg_k(\filterg(\belief,a,\bar{\physical}) )\, \filterdg(\belief,a,\bar{\physical})    \geq
\sum_{y} \uvalueg_k(T(\belief,y))\,\filterd(\belief,y).  $$
Therefore multiplying by $\pdf(\bar{\physical}|\physical)$ and integrating we have
\begin{multline}  \int_{\physicalstatespace} \sum_{a}  \uV_k(\filterg(\belief,a,\bar{\physical}) )\, \filterdg(\belief,a,\bar{\physical}) \pdf(\bar{\physical}|\physical) \, d\bar{\physical}  \\ \geq
\sum_{y} \uV_k(T(\belief,y))\,\filterd(\belief,y).  \label{eq:bdstep1}
\end{multline}

The proof of Theorem~\ref{thm:blackwell} then follows by mathematical induction using the value iteration algorithm~(\ref{eq:vi}).
Assume $\valueg_k(\belief,\physical) \geq \uvalueg_k(\belief)$ for $\belief \in  \I$.
Then 
\begin{align*}\Cost(\belief,2) + \int_{\physicalstatespace}  \sum_a \valueg_k(\filterg(\belief,a,\bar{\physical}),\bar{\physical}) \filterdg(\belief,a,\bar{\physical}) \, \pdf(\bar{\physical}|\physical) d\bar{\physical} \\  \geq
  \Cost(\belief,2) + \int_{\physicalstatespace}  \sum_a \uvalueg_k(\filterg(\belief,a,\bar{\physical}))\, \filterdg(\belief,a,\bar{\physical}) \, \pdf(\bar{\physical}|\physical) d\bar{\physical}
  \\   \geq \Cost(\belief,2) + \sum_y \uvalueg_k(\filter(\belief,\obs))\, \filterd(\belief,\obs)  
 \end{align*}
where the second inequality follows from (\ref{eq:bdstep1}).
Thus  $\valueg_{k+1}(\belief,\physical) \geq \uvalueg_{k+1}(\belief)$. This completes the induction step. Since value iteration converges pointwise,
$\valueg(\belief,\physical) \geq \uvalueg(\belief)$ thus proving the theorem.

\end{proof}

\bibliographystyle{IEEEtran}
\bibliography{$HOME/texstuff/styles/bib/vkm}

\end{document}